\documentclass[10pt,twocolumn,letterpaper]{article}

\usepackage[pagenumbers]{cvpr}

\definecolor{cvprblue}{rgb}{0.21,0.49,0.74}
\usepackage[
    pagebackref,
    breaklinks,
    colorlinks,
    allcolors=cvprblue
]{hyperref}

\def\confName{Preprint}
\def\confYear{2026}

\title{
    Hilbert States for Decision-Making under Partial Observability
}

\author{
Zibo Meng$^{1}$ \qquad
Andrew C. M. Austin$^{1}$ \qquad
Alin Achim$^{1}$\\
$^{1}$University of Bristol, United Kingdom
}

\usepackage{amsthm}

\usepackage{float}
\usepackage{placeins}
\usepackage{stfloats}
\usepackage{tikz}

\newtheoremstyle{cvprplain}
  {4pt}
  {4pt}
  {\normalfont}
  {}
  {\normalfont}
  {.}
  {0.4em}
  {}

\theoremstyle{cvprplain}

\begin{document}

\maketitle


\begin{abstract}
Decision-making under partial observability can be fundamentally limited by information, even with known dynamics: distinct physical states can induce the same observation history while favoring different actions. We study three questions: when is full-state optimality unattainable, what information exactly preserves the best observation-based decision, and how does finite-data learning affect reliability? First, we characterize decision ambiguity through conditional action preferences. We give a zero-regret criterion, a positive lower
bound for conflicting uniquely optimal actions on
positive-probability regions with uniformly positive
cost gaps, and an exact binary-action regret formula. This also shows why physical-state ambiguity alone need not cause decision loss. Second, for two actions and squared terminal-goal costs in the plane, we derive an exact inner-product representation of conditional expected cost differences using three conditional moment contrasts and a goal-dependent feature. This Hilbert state preserves observation-optimal actions
without hidden-state reconstruction. Under nondegenerate
history and goal variation, its three-dimensional form
is minimal among exact bilinear cost-contrast
representations. Third, we relate finite-data state errors to excess decision regret via goal-projected errors, action margins, and goal geometry. Physics-based simulations verify information-limited
regret, exact action preservation, and the role of
second moments. Held-out real-SAR experiments
demonstrate feasible goal-conditioned decisions
from masked images, while finite-data evaluations
reveal goal-dependent reliability.
Together, these results separate irreducible
information loss, exact representation sufficiency,
and learning-induced excess regret.
\end{abstract}


\section{Introduction}

\begin{figure*}[t]
    \centering
    \includegraphics[width=0.96\textwidth]
    {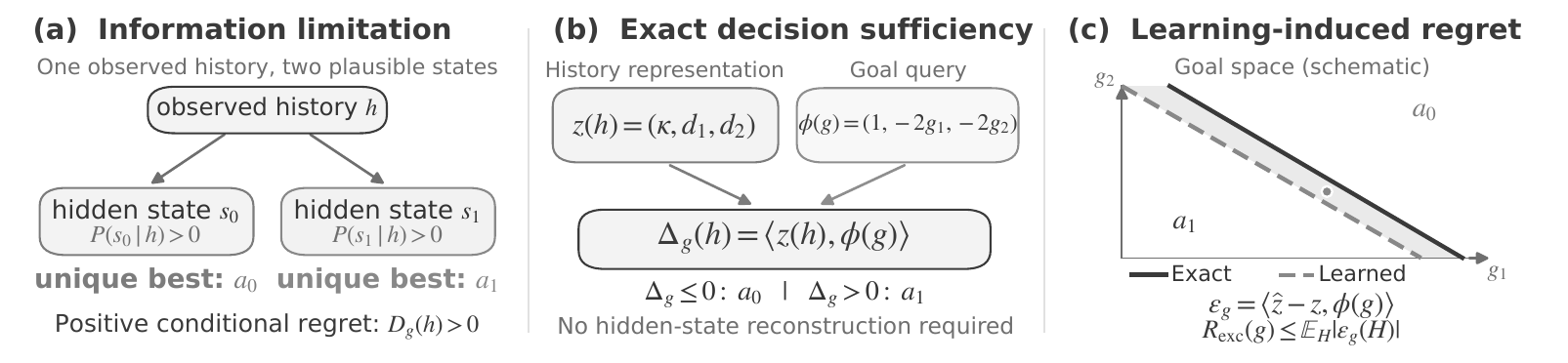}
    \caption{
        \textbf{Decision identifiability, exact
        sufficiency, and learning reliability.}
        (a) Observation-compatible hidden states
        can require opposite uniquely optimal
        actions, producing irreducible regret.
        (b) A three-dimensional Hilbert state
        exactly preserves goal-conditioned
        action-cost contrasts without hidden-state
        reconstruction under squared-goal costs.
        (c) State-estimation error affects
        decisions through its goal projection.
        The shaded region schematically illustrates
        possible disagreement between exact
        and learned decisions.
    }
    \label{fig:hilbert_decision_principle}
\end{figure*}


Partially observed dynamical systems can admit multiple physical states consistent with the same measurement history. For example, finitely sampled positions may leave an object's orientation or motion unresolved. If hidden states with positive conditional
probability require different uniquely optimal
actions for the same goal, no history-based
policy can attain the full-state optimum,
even when the dynamics are known. The gap between full-state and
observation-optimal performance is therefore
an information limitation, distinct from any
additional loss caused by imperfect
decision algorithms.


Yet physical uncertainty is not equivalent to decision uncertainty. Distinct hidden states may share an optimal action, and the best action permitted by the observations can be selected without complete state reconstruction. What must be preserved is the information needed to compare conditional expected action costs. We therefore examine when available observations preclude full-state optimal decisions, which history-dependent quantities exactly preserve the observation-optimal action, and how imperfect estimation of those quantities increases decision regret. This distinction connects fundamental information limitations with the reliability of learned decision representations.


Belief states \cite{KAELBLING199899},
predictive and Hilbert-space states
\cite{1036843.1036905,3023638.3023648},
and learned latent dynamics
\cite{pmlr-v97-hafner19a}
support decision-making under partial observability.
Task-relevant representations
\cite{NEURIPS2020_3bb585ea,NIPS2017_350db081,zhang2021learning}
further show that complete hidden-state
reconstruction is not always necessary.
We address a more specific question:
which decision losses are unavoidable
under observation ambiguity, which
conditional statistics exactly preserve
goal-dependent action comparisons,
and how does learning error affect regret?


We distinguish full-state, observation-optimal,
and learned-representation decisions.
Our contributions are threefold.

\noindent
\llap{\textbullet\hspace{0.4em}}%
We characterize decision identifiability under
partial observability, establishing a zero-regret
criterion, a positive regret bound for conflicting
optimal actions on positive-probability regions
with uniformly positive cost gaps, and an exact
binary-action regret formula. These results
distinguish physical ambiguity from unavoidable
decision loss.
\par

\noindent
\llap{\textbullet\hspace{0.4em}}%
For two actions and planar outcomes under squared-goal
costs, we construct a three-dimensional Hilbert state
whose goal-feature inner product exactly preserves
conditional action-cost contrasts. We characterize
goal-space decision geometry and establish minimality
within exact bilinear contrast representations
under explicit rank conditions.
\par

\noindent
\llap{\textbullet\hspace{0.4em}}%
We decompose decision loss into irreducible,
representation, and learning-induced components.
Exact contrast preservation yields zero representation
regret, while projected estimation errors and
decision margins bound excess regret.
Finite-data experiments identify regimes where
variance shrinkage improves or degrades decisions.
\par


\section{Related Work}

\subsection{Decisions under Partial Observability}

Blackwell's comparison of statistical experiments
formalizes how information structures constrain
achievable decision performance
\cite{10.1214/aoms/1177729032}.
Nayebi \cite{3844975.3845160} establishes
necessary predictive-information and belief-like
memory requirements for low average regret,
including under partial observability.
We characterize when states compatible with
the same observation history lack a common
optimal action and quantify the resulting
irreducible regret relative to full-state decisions.

\subsection{State and Goal Representations}

Value-directed POMDP compression examines which
belief-state information can be discarded while
preserving policy evaluation
\cite{NIPS2002_14ea0d5b}.
Kernel conditional embeddings
\cite{10.1145/1553374.1553497}
and Hilbert predictive states
\cite{3023638.3023648}
encode conditional distributions and predictions.
Goal-conditioned values \cite{pmlr-v37-schaul15}
and successor features \cite{NIPS2017_350db081}
exploit goal or reward structure.
For two actions, planar outcomes, and squared-goal
costs, we instead identify conditional statistics
that exactly preserve action-cost differences
for every goal.
This preserves observation-optimal actions and,
under explicit rank conditions, achieves minimal
dimension within exact bilinear contrast
representations.

\subsection{Learning and Decision Reliability}

Wang et al.~\cite{3618408.3619901} learn
reward-relevant belief representations for partially
observable reinforcement learning.
Allen et al.~\cite{NEURIPS2024_73073ccb}
use the $\lambda$-discrepancy to identify
non-Markovian representations, while
Kuo et al.~\cite{NEURIPS2025_286888e6}
study predictive coding for interpretable
belief recovery.
These studies address belief or memory learning;
we quantify the decision consequences of errors
in an exactly sufficient representation.
Specifically, goal-projected estimation errors
and decision margins bound excess regret relative
to the observation-optimal policy.
Finite-data experiments further examine when
variance shrinkage improves or degrades decisions.


\section{Methodology}


\subsection{Decision Ambiguity and Irreducible Regret}
\label{sec:decision_ambiguity}

Let $S\in\mathcal{S}_{\mathrm{phys}}$ denote the hidden
physical state and $H$ the observation history.
Assume standard Borel state and history spaces,
with a compatible Polish topology on
$\mathcal{S}_{\mathrm{phys}}$.
For $P_H$-almost every $h$, the regular conditional
distribution $\mu_h=P(S\in\cdot\mid H=h)$ defines
the observation-compatible state set
$\Gamma(h)=\operatorname{supp}(\mu_h)$.

Two states are observation-law equivalent when
$\mathcal L(H\mid S=s_1)=\mathcal L(H\mid S=s_2)$
under a fixed observation kernel. By contrast,
$\Gamma(h)$ is the posterior support for a realized
history, and support membership need not imply positive
posterior point mass. Neither notion alone implies
conflicting optimal actions.

Let $\mathcal A$ be a finite action set and $g$
an externally specified goal that provides no
hidden-state information beyond $H$.
For each $a\in\mathcal A$, let $c_g(s,a)$ denote
an integrable state-dependent action cost,
including relevant future dynamics.
Define
$c_g^\star(s)=\min_{a\in\mathcal A}c_g(s,a)$
and
$\mathcal A^\star(s,g)=
\arg\min_{a\in\mathcal A}c_g(s,a)$.
Writing
$\mathbb E_h[\cdot]=\mathbb E[\cdot\mid H=h]$,
the observation-based conditional cost is
$Q_g(a\mid h)=\mathbb E_h[c_g(S,a)]$.
For an observation-based policy $\pi(a\mid h,g)$,
define its conditional regret relative to
the full-state optimum as $r_g(\pi\mid h)$,
and the irreducible regret as
$D_g(h)=\inf_\pi r_g(\pi\mid h)$:
\begin{equation}
\begin{aligned}
r_g(\pi\mid h)
&=\sum_{a\in\mathcal A}\pi(a\mid h,g)Q_g(a\mid h)
-\mathbb E_h[c_g^\star(S)],\\
D_g(h)
&=\min_{a\in\mathcal A}Q_g(a\mid h)
-\mathbb E_h[c_g^\star(S)]\geq 0.
\end{aligned}
\label{eq:information_limited_gap}
\end{equation}
Since conditional regret is affine in the
action probabilities, its minimum is attained
by a deterministic action. Nonnegativity follows
from $c_g(s,a)\geq c_g^\star(s)$ for every action.

Because $\mathcal A$ is finite and statewise
excess costs are nonnegative, zero conditional
regret is attained exactly when some action
is optimal $\mu_h$-almost surely:
\begin{equation}
D_g(h)=0\iff \exists a\in\mathcal A:\;
a\in\mathcal A^\star(S,g)\quad \mu_h\text{-a.s.}
\label{eq:decision_identifiability}
\end{equation}
Thus, physical ambiguity need not imply
decision regret; neither unique state recovery
nor a unique optimal action is required.

Conflicting optimal actions induce strictly positive
irreducible regret when the associated cost gaps
are bounded away from zero.
Let $B_1,B_2\subseteq\Gamma(h)$ be disjoint
measurable sets with $p_i=\mu_h(B_i)>0$.
Assume there exist distinct actions
$a_1^\star\neq a_2^\star$ and constants
$\Delta_i>0$ such that, for $\mu_h$-almost
every $s\in B_i$ and every $a\neq a_i^\star$,
\begin{equation*}
c_g(s,a)-c_g(s,a_i^\star)\geq\Delta_i.
\end{equation*}
For any observation-based policy, write
$q_i=\pi(a_i^\star\mid h,g)$ and
$w_i=p_i\Delta_i$.
Since $q_1+q_2\leq 1$,
\begin{equation}
r_g(\pi\mid h)\ge
{\textstyle\sum_{i=1}^{2}}w_i(1-q_i)
\ge\min\{w_1,w_2\}.
\label{eq:irreducible_regret_bound}
\end{equation}
Consequently,
$D_g(h)\geq\min\{p_1\Delta_1,p_2\Delta_2\}>0$.
This bound is tight: consider two states with
probabilities $p_1+p_2=1$ and two actions
with state-dependent cost vectors
$(0,\Delta_1)$ and $(\Delta_2,0)$.
The full-state optimal cost is zero, whereas
the minimum observation-based cost equals
$\min\{p_1\Delta_1,p_2\Delta_2\}$.

For binary actions
$\mathcal A=\{a_0,a_1\}$, a stronger
exact characterization is available.
Let
$\delta_g(s)=c_g(s,a_0)-c_g(s,a_1)$.
Using
$\min\{u,v\}=(u+v-|u-v|)/2$, we obtain
\begin{equation}
\begin{aligned}
D_g(h)=\frac12\Big(
\mathbb E_h[|\delta_g(S)|]
-
|\mathbb E_h[\delta_g(S)]|
\Big).
\end{aligned}
\label{eq:exact_binary_regret}
\end{equation}
In particular,
\begin{equation}
\begin{aligned}
D_g(h)>0\iff
\mu_h(\delta_g>0)\mu_h(\delta_g<0)>0.
\end{aligned}
\label{eq:binary_conflict_criterion}
\end{equation}
Thus, irreducible binary decision loss occurs
precisely when both action preferences have
positive conditional probability.
Equivalently, $D_g(h)$ measures the failure
of state-wise cost minimization and conditional
expectation to commute.

Let $\Pi_{\mathrm{obs}}$ denote the class of
measurable observation-based policies.
For $\pi\in\Pi_{\mathrm{obs}}$, let
$A\sim\pi(\cdot\mid H,g)$ denote the selected
action, and define
$R_g(\pi)=\mathbb E[c_g(S,A)]$.
The observation-optimal risk is
$R_{\mathrm{obs}}^\star(g)
=\inf_{\pi\in\Pi_{\mathrm{obs}}}R_g(\pi)$.
The two risk benchmarks satisfy
\begin{equation}
\begin{aligned}
R_{\mathrm{full}}^\star(g)
&=\mathbb E[c_g^\star(S)],\\
R_{\mathrm{obs}}^\star(g)
&=\mathbb E_H\!\left[
\min_{a\in\mathcal A}Q_g(a\mid H)
\right].
\end{aligned}
\label{eq:optimal_risk_benchmarks}
\end{equation}
The global irreducible regret is
\begin{equation}
\begin{aligned}
R_{\mathrm{irr}}(g)
&=R_{\mathrm{obs}}^\star(g)
-R_{\mathrm{full}}^\star(g)\\
&=\mathbb E_H[D_g(H)]\geq0.
\end{aligned}
\label{eq:global_irreducible_regret}
\end{equation}
It is strictly positive when the
conflicting-action conditions above hold
on a measurable set of histories with
positive probability. This loss is imposed
by incomplete information and cannot be
eliminated by learning a more accurate
representation.

We next identify the conditional information
sufficient to attain
$R_{\mathrm{obs}}^\star(g)$
without reconstructing the hidden state.


\subsection{Hilbert Decision Sufficiency}
\label{sec:hilbert_decision_sufficiency}

Consider two admissible actions
$\mathcal A=\{a_0,a_1\}$.
Let $Y_a\in\mathbb R^2$ denote the potential
terminal outcome under action $a$, with finite
second moments.
For a specified goal
$g\in\mathcal G\subseteq\mathbb R^2$,
define the squared-goal state cost
\begin{equation}
c_g(s,a)=
\mathbb E[
    \|Y_a-g\|_2^2\mid S=s
].
\label{eq:goal_cost}
\end{equation}
We assume $Y_a\perp H\mid S$ for each
$a\in\mathcal A$, where $S$ includes all
decision-relevant physical variables and
known parameters affecting the action-conditioned
future. Conditional independence and the tower
property then give
\begin{equation}
\begin{aligned}
Q_g(a\mid h)
&=\mathbb E[c_g(S,a)\mid H=h]\\
&=\mathbb E[\|Y_a-g\|_2^2\mid H=h].
\end{aligned}
\label{eq:squared_goal_conditional_cost}
\end{equation}

For each action, define its conditional mean and second moment:
\begin{equation}
\begin{aligned}
m_a(h)&=\mathbb E[Y_a\mid H=h],\\
v_a(h)&=\mathbb E[\|Y_a\|_2^2\mid H=h].
\end{aligned}
\label{eq:conditional_moments}
\end{equation}
Conditional means alone need not preserve
the optimal action because squared-goal
risks also depend on action-dependent
conditional variance.

Define the second-moment contrast
$\kappa(h)=v_{a_0}(h)-v_{a_1}(h)$
and mean contrast
$d(h)=m_{a_0}(h)-m_{a_1}(h)\in\mathbb R^2$.
Let
$\Delta_g(h)=Q_g(a_0\mid h)-Q_g(a_1\mid h)$.
Expanding the squared-goal risks and
cancelling their common $\|g\|_2^2$
term gives
\begin{equation}
\Delta_g(h)=\kappa(h)-2g^\top d(h).
\label{eq:conditional_cost_contrast}
\end{equation}

Thus, three scalar conditional statistics
determine the exact action-cost contrast
for every $g\in\mathcal G$.
We represent them in the real Hilbert space
$\mathcal H_D=\mathbb R^3$, equipped with
the standard inner product:
\begin{equation}
\begin{aligned}
z(h)&=(\kappa(h),d_1(h),d_2(h))^\top,\\
\phi(g)&=(1,-2g_1,-2g_2)^\top.
\end{aligned}
\label{eq:hilbert_state}
\end{equation}
For $P_H$-almost every $h$ and every
$g\in\mathcal G$, the Hilbert state satisfies
\begin{equation}
\Delta_g(h)
=\langle z(h),\phi(g)\rangle_{\mathcal H_D}.
\label{eq:exact_hilbert_identity}
\end{equation}
Thus, each goal induces a continuous linear
functional of $z(h)$ that exactly preserves
the action-cost contrast without hidden-state
reconstruction.

For $a,b\in\mathcal A$, define
$\psi_{a,b}(g)
=(\mathbb I[a=a_0]-\mathbb I[b=a_0])\phi(g)$.
The exact identity extends to every action pair:
\begin{equation}
Q_g(a\mid h)-Q_g(b\mid h)
=\langle z(h),\psi_{a,b}(g)\rangle_{\mathcal H_D}.
\label{eq:action_pair_goal_feature}
\end{equation}

The Hilbert-state policy $\pi_z^\star$
selects $a_0$ when
$\langle z(h),\phi(g)\rangle_{\mathcal H_D}
\leq 0$ and $a_1$ otherwise,
breaking ties in favor of $a_0$.
By \eqref{eq:exact_hilbert_identity},
it attains $R_{\mathrm{obs}}^\star(g)$
with zero representation regret,
while irreducible regret remains unavoidable.

For fixed $h$, the $a_0$-optimal goal region is
\begin{equation}
a_0\in\arg\min_{a\in\mathcal A}Q_g(a\mid h)
\iff 2g^\top d(h)\geq\kappa(h).
\label{eq:goal_decision_region}
\end{equation}
If $d(h)\neq 0$, its boundary is
$2g^\top d(h)=\kappa(h)$, and the
Euclidean distance to this line is
\begin{equation}
r_g^{\mathrm{goal}}(h)
=\frac{|\kappa(h)-2g^\top d(h)|}
{2\|d(h)\|_2}.
\label{eq:goal_stability_radius}
\end{equation}
For $d(h)\neq0$ and $\Delta_g(h)\neq0$,
every $g'\in\mathcal G$ satisfying
$\|g'-g\|_2<r_g^{\mathrm{goal}}(h)$
preserves the unique optimal action.
If $d(h)=0$, action preference is
independent of the goal.

For any nonempty goal family
$\mathcal G\subseteq\mathbb R^2$,
\begin{equation}
\dim\operatorname{span}
\{\phi(g):g\in\mathcal G\}
=1+\dim\operatorname{aff}(\mathcal G)
\leq 3.
\label{eq:goal_feature_dimension}
\end{equation}
Thus, affine dimensions $0,1,2$ yield
goal-feature dimensions $1,2,3$,
respectively. Minimality additionally
requires the contrast-rank conditions below.

To establish minimality among exact bilinear
contrast representations, choose histories
$h_i$ and goals $g_j\in\mathcal G$.
Let $Z_{i,:}=z(h_i)^\top$,
$\Phi_{j,:}=\phi(g_j)^\top$, and
$C_{ij}=\Delta_{g_j}(h_i)$.
By \eqref{eq:exact_hilbert_identity},
\begin{equation}
C=Z\Phi^\top,\qquad
J=I_n-\frac1n\mathbf1_n\mathbf1_n^\top.
\label{eq:contrast_matrix}
\end{equation}
Here $J$ centers histories, eliminating
goal-dependent constant offsets.

Assume
\begin{equation}
\operatorname{rank}(JZ)
=\operatorname{rank}(\Phi)=3.
\label{eq:nondegenerate_rank_conditions}
\end{equation}
This requires at least four histories
whose centered contrasts span
$\mathbb R^3$ and three non-collinear
goals. Hence $JC=(JZ)\Phi^\top$
has rank three.

Any exact $r$-dimensional bilinear
contrast representation with
history-independent goal offsets satisfies
\begin{equation}
C=UV^\top+\mathbf1_n b^\top,
\label{eq:alternative_factorization}
\end{equation}
where $U\in\mathbb R^{n\times r}$,
$V\in\mathbb R^{m\times r}$, and
$b\in\mathbb R^m$.
Since $J\mathbf1_n=0$,
\begin{equation}
JC=(JU)V^\top,\qquad
3=\operatorname{rank}(JC)\le r.
\label{eq:bilinear_lower_bound}
\end{equation}
Under \eqref{eq:nondegenerate_rank_conditions},
the three-dimensional state is minimal
among exact bilinear contrast representations,
including goal-dependent offsets, on any
domain containing these histories and goals.
This does not establish minimality among
arbitrary nonlinear representations
preserving only optimal-action labels.

Exact decision sufficiency yields
$R_{\mathrm{rep}}(g):=
R_g(\pi_z^\star)-R_{\mathrm{obs}}^\star(g)=0$.
Consequently,
\begin{equation}
R_g(\pi_z^\star)-R_{\mathrm{full}}^\star(g)
=R_{\mathrm{irr}}(g).
\label{eq:hilbert_total_regret}
\end{equation}


\subsection{State Learning and Excess Regret}
\label{sec:state_learning_excess_regret}

Under the squared-goal cost, $z(h)$
exactly preserves the observation-optimal
action, but its finite-data estimate may
reverse action comparisons. We quantify
the resulting excess regret relative to
$R_{\mathrm{obs}}^\star(g)$.

Let $\mathcal D_n=
\{(H_i,Y_{i,a_0},Y_{i,a_1})\}_{i=1}^n$
contain training histories and both
action outcomes. Conditional on $\mathcal D_n$,
risk expectations use independent
histories $H$ from the target evaluation
distribution, which also defines the
conditional moments and benchmark risks.
Paired outcomes are available in simulation.
The regret bounds apply to any estimator of
$z$; learning both action-conditioned moments
from single-action observational data requires
additional identification assumptions.

Using $\mathcal D_n$, we estimate
the conditional means $\widehat m_a(h)$
and raw second moments $\widehat v_a(h)$
for each $a\in\mathcal A$, defining
\begin{equation}
\begin{aligned}
\widehat\kappa_n(h)
&=\widehat v_{a_0}(h)
 -\widehat v_{a_1}(h),\\
\widehat d_n(h)
&=\widehat m_{a_0}(h)
 -\widehat m_{a_1}(h),\\
\widehat z_n(h)
&=
\begin{pmatrix}
\widehat\kappa_n(h)\\
\widehat d_n(h)
\end{pmatrix}
\in\mathcal H_D.
\end{aligned}
\label{eq:e3_learned_hilbert_state}
\end{equation}
Let $e_n(h)=\widehat z_n(h)-z(h)$.
At inference, $\widehat z_n(h)$ uses only
the observation history, while goals
enter through $\phi(g)$; neither hidden
states nor simulator targets are required.

For each goal $g$, define
\begin{equation}
\begin{aligned}
\widehat{\Delta}_g(h)
&=\langle\widehat z_n(h),\phi(g)\rangle_{\mathcal H_D},\\
\epsilon_g(h)
&:=\widehat{\Delta}_g(h)-\Delta_g(h)\\
&=\langle e_n(h),\phi(g)\rangle_{\mathcal H_D}.
\end{aligned}
\label{eq:e3_projected_estimation_error}
\end{equation}
The learned policy $\widehat\pi_n(h,g)$
chooses $a_0$ when $\widehat\Delta_g(h)\le 0$
and $a_1$ otherwise, with ties favoring
$a_0$. Since only the goal-projected error
$\epsilon_g(h)$ perturbs the action comparison,
the optimal action is preserved for
$\Delta_g(h)\neq 0$ whenever
\begin{equation}
|\epsilon_g(h)|<|\Delta_g(h)|
\;\Longrightarrow\;
\widehat\pi_n(h,g)=\pi_z^\star(h,g).
\label{eq:e3_action_preservation}
\end{equation}
Thus, larger action-cost margins tolerate
larger projected estimation errors.

Let
$I_g(h)=\mathbf1\{
\widehat\pi_n(h,g)\neq\pi_z^\star(h,g)
\}$.
Conditioned on $\mathcal D_n$, the exact
binary-action excess regret is
\begin{equation}
\begin{aligned}
R_{\mathrm{exc}}(g\mid\mathcal D_n)
&=R_g(\widehat\pi_n\mid\mathcal D_n)
-R_{\mathrm{obs}}^\star(g)\\
&=\mathbb E_H[
|\Delta_g(H)|I_g(H)].
\end{aligned}
\label{eq:e3_exact_excess_regret}
\end{equation}
If $\Delta_g(h)\neq 0$, action disagreement
implies $|\Delta_g(h)|\leq|\epsilon_g(h)|$;
if $\Delta_g(h)=0$, excess cost is zero.
When $\mathbb E_H[\epsilon_g(H)^2]<\infty$,
we obtain
\begin{equation}
\begin{aligned}
R_{\mathrm{exc}}(g\mid\mathcal D_n)
&\leq\mathbb E_H[|\epsilon_g(H)|]\\
&\leq\sqrt{\mathbb E_H[\epsilon_g(H)^2]}.
\end{aligned}
\label{eq:e3_margin_excess_regret_bound}
\end{equation}
These bounds link decision regret to
goal-projected estimation error and true
action margins. Lower state-estimation
error alone need not reduce regret;
no convergence rate in $n$ is established.

For a finite goal family
$\mathcal G=\{g_1,\ldots,g_k\}$
with $w_j>0$ and $\sum_jw_j=1$,
define
\begin{equation}
M_{\mathcal G}
=\sum_{j=1}^{k}
w_j\phi(g_j)\phi(g_j)^\top.
\label{eq:e3_goal_error_geometry}
\end{equation}
The single-goal bound and weighted
Cauchy--Schwarz yield
\begin{equation}
\begin{gathered}
\sum_{j=1}^{k}
w_jR_{\mathrm{exc}}(g_j\mid\mathcal D_n)
\\[-2pt]
\leq
\sqrt{
\mathbb E_H[
e_n(H)^\top M_{\mathcal G}e_n(H)]
}.
\end{gathered}
\label{eq:e3_goal_family_regret_bound}
\end{equation}
Since $w_j>0$,
$\operatorname{rank}(M_{\mathcal G})
=1+\dim\operatorname{aff}(\mathcal G)\leq 3$.
Thus, only goal-spanned estimation-error
directions enter the regret bound.
This guarantees neither reliability
for unseen goals nor regret minimization
through quadratic-error minimization.

To shrink the estimated variance contribution
to the second-moment contrast, define
$\widehat\kappa_m(h)
=\|\widehat m_{a_0}(h)\|_2^2
-\|\widehat m_{a_1}(h)\|_2^2$
and
$\widehat\tau(h)
=\widehat\kappa_n(h)-\widehat\kappa_m(h)$.
For $\lambda\in[0,1]$, the
variance-shrunk Hilbert state is
\begin{equation}
\widehat z_{n,\lambda}(h)
=
\begin{pmatrix}
\widehat\kappa_m(h)
+\lambda\widehat\tau(h)\\
\widehat d_n(h)
\end{pmatrix}.
\label{eq:e3_variance_shrunk_state}
\end{equation}
The endpoints recover the mean-only
($\lambda=0$) and estimated full-moment
($\lambda=1$) comparisons; intermediate
values trade variance-estimation noise
against shrinkage bias.

We select $\lambda$ by
leave-one-simulator-seed-out
cross-validation using training data only;
development and shifted-development
sets are reserved for evaluation.
Conditional on $\mathcal D_n$, the same
regret bounds hold for the policy induced
by $\widehat z_{n,\lambda}$, with
$e_{n,\lambda}(h)=
\widehat z_{n,\lambda}(h)-z(h)$,
including training-selected $\lambda$.

Combining irreducible and learning-induced
regret gives
\begin{equation}
\begin{gathered}
R_g(\widehat\pi_n\mid\mathcal D_n)
-R_{\mathrm{full}}^\star(g)
\\[-2pt]
=R_{\mathrm{irr}}(g)
+R_{\mathrm{exc}}(g\mid\mathcal D_n).
\end{gathered}
\label{eq:e3_total_regret_decomposition}
\end{equation}
The decomposition separates unavoidable
partial-observation loss
($R_{\mathrm{irr}}$) from finite-data
decision error ($R_{\mathrm{exc}}$).
Exact Hilbert sufficiency yields zero
representation regret.


\section{Experiments}


\subsection{Decision Limits under Partial Observability}
\label{sec:exp_decision_limits}

We construct observation-equivalent
physical states with opposite unique
full-state-optimal actions and quantify
the resulting irreducible regret.
Known dynamics and exact
observation-based optimization isolate
this loss from learning error.

We consider a dynamical system with
hidden orientation $q\in\mathbb S^2$,
observed planar positions, and two
candidate future action programs.
For goal $g\in\mathbb R^2$, the
objective is to minimize squared
terminal-position error.
The observation-based oracle receives
$H=(p_3,p_4,\chi,\eta)$,
where $\chi$ specifies the observation
context and $\eta=(\beta,\rho,F)$
contains the known simulator parameters.

\begin{samepage}

In the passive context,
$T_\beta(q)=R_z(\theta(q))q$, where
$\theta(q)=\tfrac12(1+\beta q_xq_y)$,
commutes with the reflection
$\mathcal R(x,y,z)=(x,y,-z)$.
\begin{equation}
\begin{aligned}
\mathcal H_0(q_0;\eta)
&=\mathcal H_0(\mathcal Rq_0;\eta),\\
q_1^\pm(h)
&=\left(
\frac{p_3}{F},
\pm\sqrt{1-\|p_3/F\|_2^2}
\right).
\end{aligned}
\label{eq:exp_physical_ambiguity}
\end{equation}

\end{samepage}

The observation identity follows because
$p_3,p_4$ are invariant under
$\mathcal R$.
For $\|p_3/F\|_2<1$, the two
compatible orientations are distinct.
Since both the passive observation map
and the spherical prior are
reflection-invariant, their posterior
weights conditional on $h$ satisfy
$p_+=p_-=\tfrac12$.
Thus, exact knowledge of the dynamics
does not identify the hidden orientation.

For each compatible branch $q_1^\pm(h)$,
let $\delta_g^\pm(h)$ denote the
squared terminal-position cost of
$a_0$ minus that of $a_1$.
Each contrast is affine in $g$.
The branches have opposite unique
optimal actions exactly for goals in
$\mathcal C(h)=
\{g:\delta_g^+(h)\delta_g^-(h)<0\}$;
the boundary of $\mathcal C(h)$ lies
on their action-indifference lines.

For $p_+=p_-=\tfrac12$, the exact
binary-regret identity
\eqref{eq:exact_binary_regret} reduces to
\begin{equation}
D_g(h)=\tfrac12
\min\{|\delta_g^+(h)|,|\delta_g^-(h)|\}
\mathbf1\{g\in\mathcal C(h)\}.
\label{eq:exp_exact_regret}
\end{equation}
For this symmetric two-branch posterior,
$D_g(h)>0$ exactly when the branches
have opposite unique optimal actions;
otherwise, $D_g(h)=0$.

Each development regime contains 1,536
histories, stratified into six passive
and six active groups of 128 histories
each. We evaluate every history on
25 fixed exogenous goals selected from
the training regime, obtaining 38,400
history--goal pairs per regime.

From the compatible physical branches
and action-conditioned futures, we
directly evaluate the empirical
full-state and observation-based oracle
risks, $\widehat R_{\mathrm{full}}^\star$
and $\widehat R_{\mathrm{obs}}^\star$,
without fitting a predictor.
Their difference yields the empirical
irreducible regret,
$\widehat R_{\mathrm{irr}}
=\widehat R_{\mathrm{obs}}^\star
-\widehat R_{\mathrm{full}}^\star$.

\begin{figure}[t]
    \centering
    \includegraphics[width=0.88\columnwidth]
    {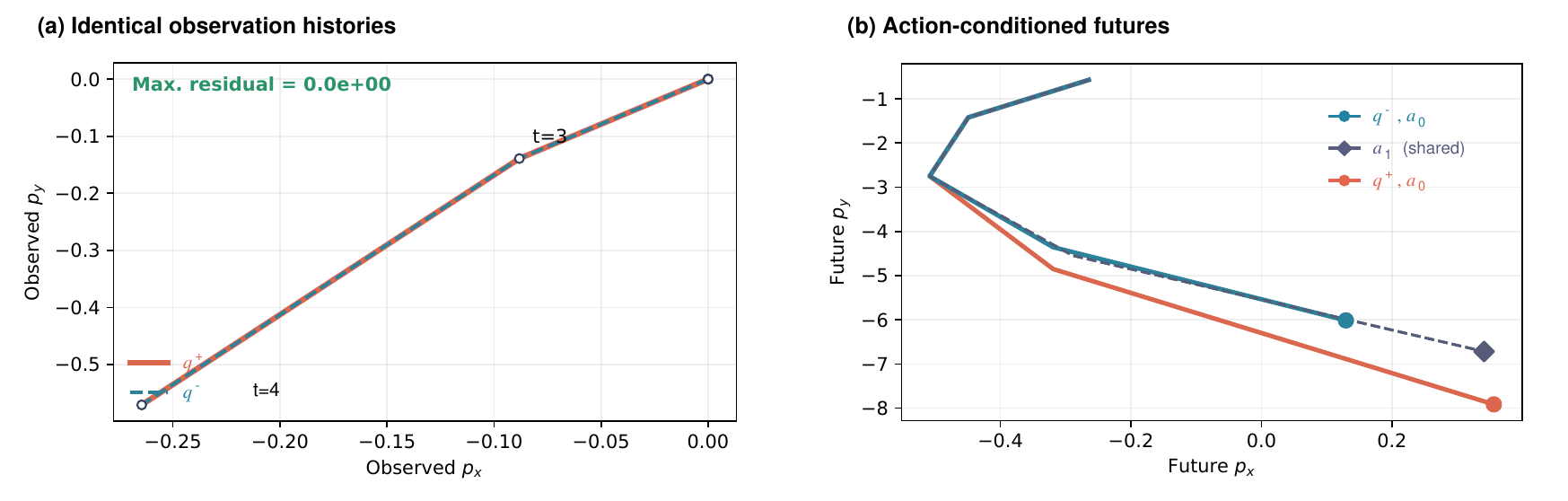}
    \caption{
    Physical-simulation witness of decision
    non-identifiability.
    (a) Distinct hidden states yield identical
    observation histories.
    (b) Their action-conditioned futures imply
    opposite optimal actions for
    $g=(1.252,8.957)$
    (irreducible Bayes regret: $10.5628$).
    }
    \label{fig:physical_ambiguity}
\end{figure}

Figure~\ref{fig:physical_ambiguity} provides a physical-model
witness of decision non-identifiability.
By balanced sampling, 50\% of the
38,400 history--goal pairs per regime
are physically ambiguous.
Conflicting optimal actions occur in
13.49\% (DEV) and 15.13\%
(\texttt{SHIFTED\_DEV}) of all pairs,
corresponding to 26.98\% and 30.27\%
of ambiguous pairs, respectively.
The empirical irreducible regrets are
0.3319 and 0.3977, matching the
certified lower bounds with maximum
discrepancy below $1.2\times10^{-13}$.
Complete diagnostics are provided
in the supplementary material.

In active contexts, an additional
first-axis rotation yields a unique
compatible orientation.
All six active groups therefore have
zero irreducible regret, whereas
all six passive groups exhibit
positive average regret.

These results establish a physical
decision limit: state ambiguity alone
need not cause decision loss, but
conflicting optimal actions impose
irreducible regret even with known
dynamics and exact observation-based
optimization.

\begin{figure*}[t]
    \centering
    \includegraphics[
        width=0.70\textwidth
    ]{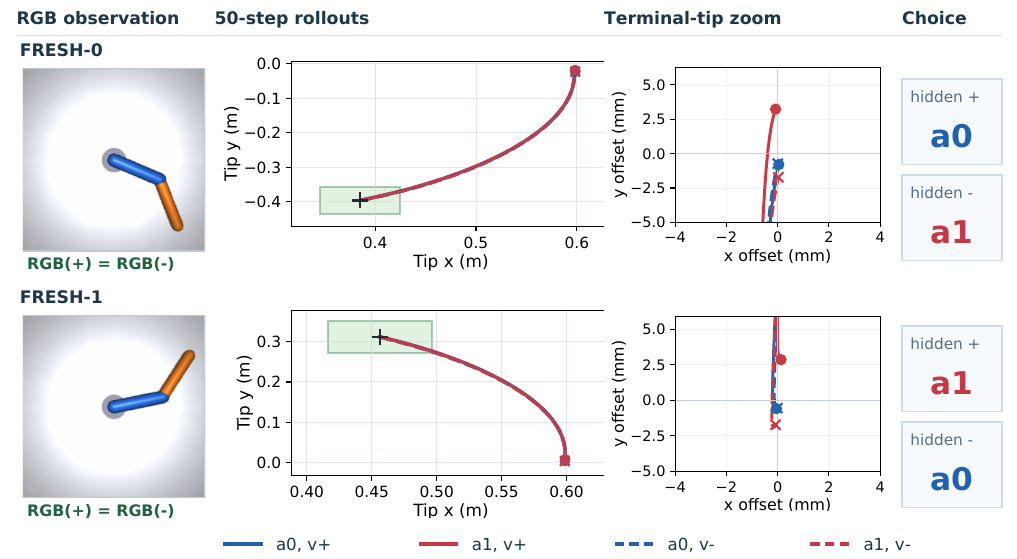}

    \caption{
    Physical evidence of decision non-identifiability.
    Two MuJoCo states exhibit pixelwise-identical RGB
    observations but different hidden velocities and
    opposite optimal actions.
    Their 50-step trajectories and terminal positions
    reveal the hidden physical differences.
    Positive action-cost gaps are certified over the
    registered goal region, with numerical evidence
    provided in the supplement.
    }

    \label{fig:mujoco_image_ambiguity}
\end{figure*}

Figure~\ref{fig:mujoco_image_ambiguity}
provides image-level counterfactual evidence:
pixelwise-identical RGB observations conceal
different physical dynamics and opposite
optimal actions.
The resulting decision conflict reflects
missing information rather than a failure
of visual feature learning.


\subsection{Exactness of Hilbert Decision States}
\label{sec:exp_hilbert_exactness}

We test whether the three-dimensional
Hilbert state preserves observation-optimal
decisions without reconstructing hidden
orientations. Using the system
of Section~\ref{sec:exp_decision_limits},
we compare the physical oracle, which
marginalizes over compatible orientations,
with the Hilbert policy using only
$z(h)$ and $\phi(g)$.
Both are evaluated on the same
38,400 history--goal pairs per regime.
Table~\ref{tab:exp_hilbert_exactness}
reports the results.

\begin{table}[t]
\centering
\caption{
Exact Hilbert decision certification.
Each regime contains 1,536 histories
and 25 goals.
}
\label{tab:exp_hilbert_exactness}
\small
\setlength{\tabcolsep}{2.5pt}
\renewcommand{\arraystretch}{1.04}
\begin{tabular}{@{}lcc@{}}
\toprule
Quantity & Dev & Shifted dev \\
\midrule
Max. identity discrepancy
& $5.83\times10^{-13}$ & $7.25\times10^{-13}$ \\
Hilbert action agreement (\%)
& 100.00 & 100.00 \\
Ranks of $JZ,\Phi,JC$
& $3/3/3$ & $3/3/3$ \\
Mean-only disagreement (\%)
& 0.61 & 0.49 \\
Mean-only excess regret
& 0.00280 & 0.00206 \\
Hilbert representation regret
& 0.00000 & 0.00000 \\
\bottomrule
\end{tabular}
\end{table}

Let $\mathcal B(h)$ index the compatible
physical orientations
$\Gamma(h)=\{q^b(h):b\in\mathcal B(h)\}$,
with posterior probabilities $p_b(h)$
and branch-specific action-cost contrasts
$\delta_g^b(h)$.
The physical conditional contrast
has the exact representation
\begin{equation}
\begin{aligned}
\Delta_g^{\mathrm{phys}}(h)
&=\sum_{b\in\mathcal B(h)}
p_b(h)\delta_g^b(h)\\
&=\Delta_g^\mu(h)+\Delta^\Sigma(h)\\
&=\kappa(h)-2g^\top d(h)
=\langle z(h),\phi(g)\rangle,
\end{aligned}
\label{eq:exp_hilbert_identity}
\end{equation}
where
$\Delta_g^\mu=
\|m_{a_0}-g\|_2^2-\|m_{a_1}-g\|_2^2$
and
$\Delta^\Sigma=
\operatorname{tr}(\Sigma_{a_0}-\Sigma_{a_1})$.
Thus, the exact action contrasts
depend on relative conditional means
and a second-moment correction,
without full-state recovery.
Table~\ref{tab:exp_hilbert_exactness}
certifies numerical agreement;
detailed residual checks are
deferred to the appendix.

By \eqref{eq:goal_decision_region},
the Hilbert policy selects $a_0$
precisely when
$2g^\top d(h)\geq\kappa(h)$,
with ties favoring $a_0$.
The physical and Hilbert oracles agree
on all 38,400 history--goal pairs
in each regime
(Table~\ref{tab:exp_hilbert_exactness}).
For every evaluated history,
$d(h)\neq0$, so the decision boundary
is a line in goal space.
Stability-radius and boundary-crossing
formulas are derived in the appendix.

For non-collinear planar goals,
the goal-feature span has dimension
three by \eqref{eq:goal_feature_dimension}.
In both evaluation regimes,
Table~\ref{tab:exp_hilbert_exactness}
reports
$\operatorname{rank}(JZ)
=\operatorname{rank}(\Phi)
=\operatorname{rank}(JC)=3$.
Under the exact full-rank conditions in \eqref{eq:nondegenerate_rank_conditions},
\eqref{eq:bilinear_lower_bound} establishes three-dimensional minimality
within exact bilinear cost-contrast representations,
even with goal-dependent offsets shared across
histories. The observed numerical ranks are
consistent with these conditions, but do not
independently certify exact algebraic rank without
verified approximation-error bounds.

\begin{figure*}[!t]
    \centering
    \includegraphics[width=0.82\textwidth]
    {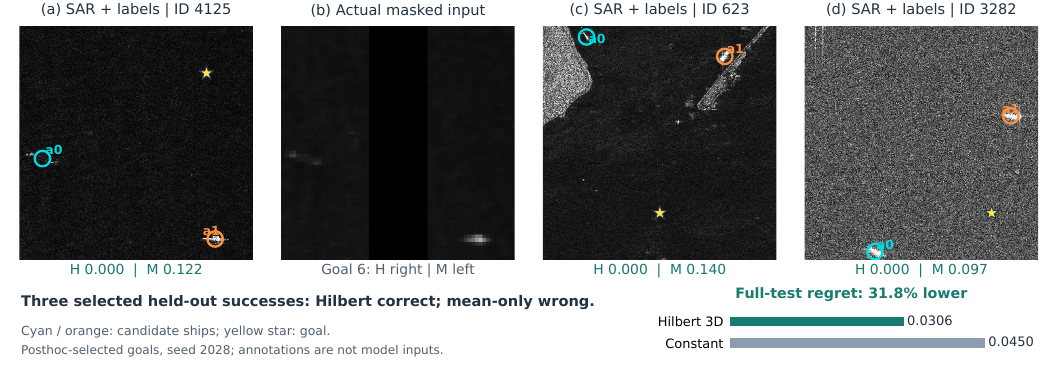}
    \caption{
        Decision preservation on real SAR images.
        (a,c,d) Three post-hoc selected HRSID examples
        with candidate actions (cyan/orange) and goals (yellow).
        (b) The corresponding masked observation.
        H/M denote Hilbert/mean-only regret relative
        to annotation-derived optimal actions;
        Hilbert matches this oracle in all shown cases.
        Inset: full-test regret of $0.0306$ versus $0.0450$
        for the constant-state baseline ($31.8\%$ lower).
        Annotations are not model inputs.
    }
    \label{fig:real_sar_decision}
\end{figure*}

To isolate the second-moment contribution,
we compare against the mean-only oracle
$\pi_\mu(h,g)=\arg\min_{a\in\mathcal A}
\|m_a(h)-g\|_2^2$, with ties favoring $a_0$.
As Table~\ref{tab:exp_hilbert_exactness}
shows, mean-only disagrees with the exact
Hilbert policy on $0.61\%$ (Dev) and
$0.49\%$ (Shifted dev) of history--goal
pairs, incurring excess regrets of
$0.00280$ and $0.00206$, respectively.
All disagreements occur in passive
contexts with nonzero action-variance
contrasts; the policies agree in evaluated
active contexts, where $\Delta^\Sigma(h)=0$.
A complementary two-model construction
in the supplementary material shows why
identical conditional means need not
preserve optimal actions.

The zero representation regret in
Table~\ref{tab:exp_hilbert_exactness}
implies, by \eqref{eq:hilbert_total_regret},
that full-state-relative regret is entirely
irreducible: $0.33191$ (Dev) and
$0.39769$ (Shifted dev).
Thus, observation-optimal decisions
require no hidden-state recovery.
We next examine finite-data estimation
of $z(h)$.


\subsection{Learning and Decision Reliability}
\label{sec:learning_reliability}

\begin{figure*}[!t]
    \centering
    \includegraphics[width=0.90\textwidth]
    {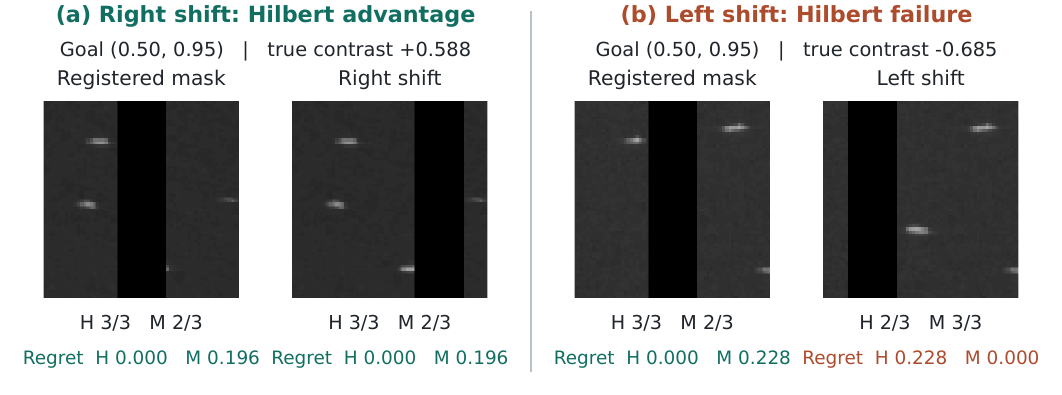}
    \caption{
    Real-SAR decision witnesses under observation-and-goal
    OOD. Two selected HRSID images at an unseen goal
    $g=(0.50,0.95)$ compare registered and shifted masks.
    H/M denote Hilbert-3D/mean-only; correctness is measured
    over three seeds, and mean regret is relative to
    annotation-derived optimal actions. Hilbert's advantage
    persists under a rightward shift (a) but reverses under
    a leftward shift (b). A fixed gamma correction
    ($\gamma=0.55$) is applied for visualization only;
    model inputs and predictions are unchanged.
    These witnesses do not imply uniform OOD superiority.
    }
    \label{fig:main_ood}
\end{figure*}

We evaluate learned Hilbert states under finite data
and distribution shift, comparing mean-only,
full-moment, and variance-shrunk estimators.
Leave-one-simulator-seed-out cross-validation on
25 registered goals selects $\lambda_{\mathrm{CV}}=0.4094$
at $n=6912$. Relative to full-moment estimation,
CV shrinkage reduces excess regret by $37.4\%$
(DEV) and $40.7\%$ (SHIFTED\_DEV), although mean-only
performs better than shrinkage on SHIFTED\_DEV.

Across 168 configurations, the exact excess-regret
identity, projected-error bounds, and regret
decomposition are numerically verified without
violations (maximum decomposition residual below
$4\times10^{-13}$). Projected-error ranks are three
for 25 extrapolated goals and two for the affine-line
families, consistent with the theoretical rank bound.
On extrapolated DEV goals, shrinkage reduces regret
from $0.13680$ to $0.08931$ (95\% paired CI for the
difference: $[-0.08138,-0.01607]$).
Conversely, on novel-direction SHIFTED\_DEV goals,
regret increases from $0.06181$ to $0.08709$
(95\% paired CI: $[0.00517,0.04854]$).
Thus, exact decision sufficiency does not guarantee
uniform finite-data improvements. Full estimator
comparisons, regret decompositions, and numerical
certificates are detailed in the supplementary material.

We evaluate learned decision states on partially
masked $64\times64$ HRSID SAR observations,
comparing two image-region actions across nine
spatial goals. The source-prefix-disjoint partition
contains 1,404 training, 256 validation, and
321 held-out test images. Bounding-box annotations
define action outcomes and oracle decisions but
are not provided as model inputs.

Across three training seeds, Hilbert 3D achieves
oracle-relative regret of $0.03064\pm0.00133$,
versus $0.04495$ for the constant-state baseline
($31.8\%$ lower), $0.03348\pm0.00994$ for
mean-only estimation, and $0.02863\pm0.00058$
for direct nine-goal contrast prediction.
Exploratory bootstrap resampling over 20 held-out
filename-prefix groups yields a Hilbert-minus-constant
difference of $-0.01431$ (95\% percentile interval:
$[-0.01850,-0.00932]$); these groups do not establish
independent SAR acquisitions.
Figure~\ref{fig:real_sar_decision} shows selected
held-out decisions and their masked observations.
The improvement over mean-only is not statistically
resolved, while direct contrast achieves slightly
lower average regret on the evaluated goals.

We evaluate eight unseen goals under five observation
masks on the same 321 held-out HRSID images.
For unseen goals, predicted action contrasts are
obtained through a least-squares affine extension
of nine registered-goal predictions, without
using ground-truth contrasts in fitting.
Changing the observation mask alters the available
image pixels while preserving the underlying image,
goal, and annotation-defined action outcomes.
Figure~\ref{fig:main_ood} shows two selected witnesses.
Under a rightward shift, Hilbert-3D remains correct
across all three seeds, versus two for mean-only.
Under a leftward shift, Hilbert correctness falls
from three to two seeds (regret $0.228$), while
mean-only incurs zero regret.
Across the full evaluation, Hilbert/mean-only
regrets are $0.04025/0.04296$ (registered),
$0.05611/0.06155$ (right shift), and
$0.08501/0.06494$ (left shift).
These results connect exact decision sufficiency
to practical goal-conditioned decisions without
full hidden-state reconstruction, while showing
that finite-data approximations need not retain
uniform reliability under observation shifts.

Additional real-robot experiments in the
supplementary material demonstrate
goal-conditioned outcome forecasting
without goal-specific retraining.
These offline results do not establish
causal action regret or deployed-policy
performance.


\section{Conclusion}

We establish when partial observability
causes irreducible decision regret and
show that a three-dimensional Hilbert
state exactly preserves observation-optimal
actions under squared-goal costs without
hidden-state reconstruction.
Under the stated full-rank conditions,
three dimensions are minimal among exact
bilinear cost-contrast representations,
even with history-independent goal offsets.
Physical simulations verify decision
non-identifiability and zero representation
regret, while held-out real-SAR experiments
demonstrate goal-conditioned decision learning
from incomplete images.
Recorded real-robot demonstrations further
support reusable goal-conditioned outcome
forecasting without goal-specific retraining.
Finite-data results characterize how
estimation error and goal geometry affect
decision reliability.

\section*{LLM Usage Statement}
OpenAI's ChatGPT was used to edit or refine author-produced writing, related work and citations, code and implementation, and experimental design, execution, and results analysis. All scientific claims, theoretical statements, experimental designs, implementations, numerical results, and final manuscript content were reviewed and verified by the authors. The authors take full responsibility for the content of this paper.

\bibliographystyle{ieeenat_fullname}
\bibliography{main}

\clearpage
\numberwithin{equation}{section}
\numberwithin{figure}{section}
\numberwithin{table}{section}

\section{Conditional-Moment Expansion}
\label{app:moment_cost_expansion}

For each action $a\in\mathcal A$,
let
$m_a(h)=\mathbb E[Y_a\mid H=h]$
and
$v_a(h)=\mathbb E[\|Y_a\|_2^2\mid H=h]$.
Define
$\Sigma_a(h)=\operatorname{Cov}(Y_a\mid H=h)$.

Expanding the squared-goal cost yields
\begin{equation}
\begin{aligned}
Q_g(a\mid h)
&=v_a(h)-2g^\top m_a(h)+\|g\|_2^2\\
&=\|m_a(h)-g\|_2^2
  +\operatorname{tr}\Sigma_a(h).
\end{aligned}
\label{eq:moment_cost_identity}
\end{equation}

The second equality follows from
\begin{equation}
v_a(h)
=\|m_a(h)\|_2^2
+\operatorname{tr}\Sigma_a(h).
\end{equation}

We next establish simultaneous validity
over the entire goal family.
Under $Y_a\perp H\mid S$ and finite second moments,
the conditional tower property yields
\begin{equation}
\label{eq:app_conditional_tower}
\begin{aligned}
Q_g(a\mid h)
&=\mathbb E_h\!
  \left[\mathbb E[\|Y_a-g\|_2^2\mid S]\right]\\
&=\mathbb E[\|Y_a-g\|_2^2\mid H=h].
\end{aligned}
\end{equation}
The conditional moments $m_a(h)$ and $v_a(h)$
admit common versions outside a single
$P_H$-null set. Since the squared-goal expansion
is a polynomial identity in $g$ with these
fixed coefficients, it holds simultaneously
for every $g\in\mathcal G$ on that set.
No goal-dependent exceptional set is required.

Subtracting the conditional risks of
$a_0$ and $a_1$ cancels the common
$\|g\|_2^2$ term and gives
$\Delta_g(h)=\kappa(h)-2g^\top d(h)$.

To derive the variance decomposition,
subtract the second-moment identities
for the two actions:
\begin{equation}
\begin{aligned}
v_{a_0}(h)-v_{a_1}(h)
&=\|m_{a_0}(h)\|_2^2
 -\|m_{a_1}(h)\|_2^2\\
&\quad+
\operatorname{tr}\bigl(
\Sigma_{a_0}(h)-\Sigma_{a_1}(h)
\bigr)\\
&=\kappa_m(h)+\tau(h).
\end{aligned}
\label{eq:app_variance_decomposition_derivation}
\end{equation}
This establishes the mean--variance
decomposition of the second-moment
contrast. The correction $\tau(h)$
equals the difference between the
conditional covariance traces
of the two actions.

Conditional means alone need not determine
the optimal action.
Consider a constant observation history $H=h$,
a Rademacher random variable
$U\in\{-1,1\}$ with equal probabilities,
and a unit vector $u\in\mathbb R^2$.
Construct two admissible models:
\begin{align*}
\text{Model I:}\quad
&(Y_{a_0},Y_{a_1})=(Uu,0),\\
\text{Model II:}\quad
&(Y_{a_0},Y_{a_1})=(0,Uu).
\end{align*}
In both models,
$m_{a_0}(h)=m_{a_1}(h)=0$.
However, their second-moment contrasts satisfy
$\kappa_{\mathrm I}(h)=1$ and
$\kappa_{\mathrm{II}}(h)=-1$.
Consequently, for every $g\in\mathbb R^2$,
\begin{equation*}
\Delta_g^{\mathrm I}(h)=1,
\qquad
\Delta_g^{\mathrm{II}}(h)=-1.
\end{equation*}
Thus, the unique optimal actions are
$a_1$ in Model I and $a_0$ in Model II.
Because the conditional action means and
the goal are identical in both models,
no decision rule based exclusively on
these means can preserve the optimal
action uniformly over the two models.

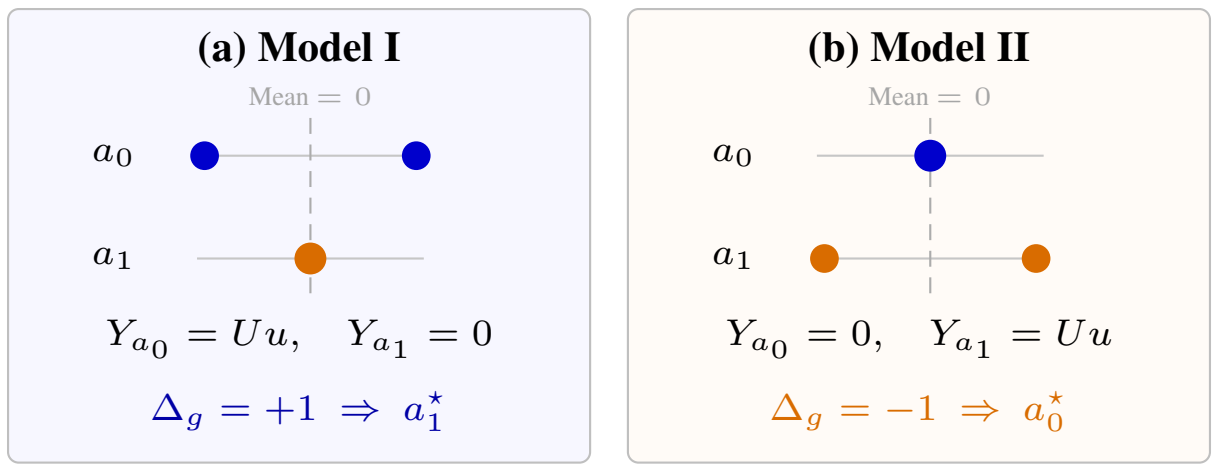
\begin{figure}[t]
\centering

\begin{tikzpicture}[
    x=1cm,
    y=1cm,
    font=\scriptsize
]


\draw[
    rounded corners=2pt,
    draw=gray!50,
    fill=blue!3
]
(0.05,0) rectangle (3.90,3.0);

\node[
    font=\bfseries\footnotesize
] at (1.975,2.73)
{(a) Model I};

\node at (0.75,2.03) {$a_0$};
\node at (0.75,1.35) {$a_1$};

\draw[gray!45]
(1.30,2.03) -- (2.80,2.03);

\draw[gray!45]
(1.30,1.35) -- (2.80,1.35);

\draw[
    densely dashed,
    gray!65
]
(2.05,1.12) -- (2.05,2.30);

\fill[blue!80!black]
(1.35,2.03) circle (2.7pt);

\fill[blue!80!black]
(2.75,2.03) circle (2.7pt);

\fill[orange!85!black]
(2.05,1.35) circle (3pt);

\node[
    font=\tiny,
    gray!70
] at (2.05,2.42)
{Mean $=0$};

\node at (1.975,0.82)
{$Y_{a_0}=Uu,\quad Y_{a_1}=0$};

\node[
    font=\bfseries\scriptsize,
    text=blue!65!black
] at (1.975,0.34)
{$\Delta_g=+1\ \Rightarrow\ a_1^\star$};


\draw[
    rounded corners=2pt,
    draw=gray!50,
    fill=orange!3
]
(4.15,0) rectangle (8.00,3.0);

\node[
    font=\bfseries\footnotesize
] at (6.075,2.73)
{(b) Model II};

\node at (4.85,2.03) {$a_0$};
\node at (4.85,1.35) {$a_1$};

\draw[gray!45]
(5.40,2.03) -- (6.90,2.03);

\draw[gray!45]
(5.40,1.35) -- (6.90,1.35);

\draw[
    densely dashed,
    gray!65
]
(6.15,1.12) -- (6.15,2.30);

\fill[blue!80!black]
(6.15,2.03) circle (3pt);

\fill[orange!85!black]
(5.45,1.35) circle (2.7pt);

\fill[orange!85!black]
(6.85,1.35) circle (2.7pt);

\node[
    font=\tiny,
    gray!70
] at (6.15,2.42)
{Mean $=0$};

\node at (6.075,0.82)
{$Y_{a_0}=0,\quad Y_{a_1}=Uu$};

\node[
    font=\bfseries\scriptsize,
    text=orange!85!black
] at (6.075,0.34)
{$\Delta_g=-1\ \Rightarrow\ a_0^\star$};

\end{tikzpicture}

\caption{
Why conditional means are insufficient
for optimal action selection.
Both models share $H=h$ and
$m_{a_0}=m_{a_1}=0$.
In (a), $Y_{a_0}=Uu$ and $Y_{a_1}=0$;
(b) interchanges these outcomes,
where $U=\pm1$ with equal probability
and $\|u\|_2=1$.
Under squared-goal costs, the exact
contrasts are $\Delta_g=+1$ and $-1$
for every goal $g$, respectively,
yielding opposite optimal actions.
The illustration is schematic.
}

\label{fig:variance_moment_necessity}

\end{figure}

\section{Binary-Branch Regret Derivation}
\label{app:binary_branch_regret}

We first derive the general binary-action
irreducible-regret identity.
Fix a goal $g$ and a history $h$ for which
the conditional action costs are integrable.
Write
$\delta=\delta_g(S)=c_g(S,a_0)-c_g(S,a_1)$,
and define
$\delta_+=\max\{\delta,0\}$ and
$\delta_-=\max\{-\delta,0\}$.

Using the identity
$\min\{x,y\}=(x+y-|x-y|)/2$
in the definition of conditional regret gives
\begin{equation}
\label{eq:app_general_binary_regret}
\begin{aligned}
D_g(h)
&=\frac12\left(
\mathbb E_h|\delta|
-|\mathbb E_h\delta|
\right)\\
&=\min\left\{
\mathbb E_h\delta_+,
\mathbb E_h\delta_-
\right\}.
\end{aligned}
\end{equation}
Since a nonnegative integrable random variable
has zero expectation if and only if it
vanishes almost surely, we obtain
\begin{equation}
\label{eq:app_zero_binary_regret}
D_g(h)=0
\iff
\begin{cases}
\mu_h(\delta>0)=0,\quad\text{or}\\
\mu_h(\delta<0)=0.
\end{cases}
\end{equation}
Therefore, positive posterior probability
on both strict signs is necessary and
sufficient for strictly positive binary-action
irreducible regret. No uniform cost-gap
assumption is needed for this qualitative
characterization. States satisfying
$\delta=0$ may have arbitrary posterior
probability without changing the criterion.

Consider two observation-compatible
physical branches with equal posterior
probabilities $p_+=p_-=\tfrac12$.
For each branch $s\in\{+,-\}$,
let $c_a^s$ denote the terminal cost
under action $a$, and define
$\delta_g^s=c_{a_0}^s-c_{a_1}^s$.

The conditional irreducible regret is
the difference between the optimal
observation-based cost and the
full-state optimal cost:
\begin{equation}
\begin{aligned}
D_g(h)
&=\min_a\frac{c_a^++c_a^-}{2}\\
&\quad-\frac12
\left(\min_a c_a^+
+\min_a c_a^-\right).
\end{aligned}
\end{equation}

Using
$\min\{x,y\}=(x+y-|x-y|)/2$,
we obtain
\begin{equation}
\begin{aligned}
D_g(h)
&=\frac14\left(
|\delta_g^+|+|\delta_g^-|
-|\delta_g^++\delta_g^-|
\right)\\
&=\frac12
\min\{|\delta_g^+|,|\delta_g^-|\}
\mathbf1\{
\delta_g^+\delta_g^-<0
\}.
\end{aligned}
\label{eq:app_binary_branch_regret}
\end{equation}

The two-branch irreducible regret is
strictly positive exactly when
$\delta_g^+\delta_g^-<0$.
If $\delta_g^+\delta_g^-\geq 0$,
the two branches admit a common
statewise optimal action, including
when either branch has tied optimal
actions, and $D_g(h)=0$.

Importantly, a tie between the
observation-conditioned action costs
does not generally eliminate
irreducible regret.
When $\delta_g^+=-\delta_g^-\neq 0$,
we have
$Q_g(a_0\mid h)=Q_g(a_1\mid h)$,
but
$D_g(h)=|\delta_g^+|/2>0$.

We next establish a sharp lower bound
when two posterior regions favor
distinct optimal actions.
Let $B_1,B_2\subseteq\Gamma(h)$ be disjoint
measurable sets with
$p_i=\mu_h(B_i)>0$.
Suppose distinct actions
$a_1^\star\neq a_2^\star$ and constants
$\Delta_i>0$ satisfy
\[
c_g(s,a)-c_g(s,a_i^\star)\geq\Delta_i
\]
for $\mu_h$-almost every $s\in B_i$
and every $a\neq a_i^\star$.
Then
\begin{equation}
\label{eq:app_conflict_lower_bound}
D_g(h)\geq
\min\{p_1\Delta_1,p_2\Delta_2\}>0.
\end{equation}

To establish the bound, write
$c_g^\star(s)=\min_{a\in\mathcal A}c_g(s,a)$.
For any fixed action $a$, the inequality
$a_1^\star\neq a_2^\star$ guarantees
that $a\neq a_i^\star$ for at least
one $i\in\{1,2\}$.
Consequently,
\begin{equation}
\label{eq:app_conflict_proof}
\begin{aligned}
&\mathbb E_h[
c_g(S,a)-c_g^\star(S)]\\
&\quad\geq p_i\Delta_i\\
&\quad\geq
\min\{p_1\Delta_1,p_2\Delta_2\}.
\end{aligned}
\end{equation}
Minimizing over $a$ proves the claim.
Randomizing over actions cannot improve
the bound because conditional expected
cost is affine in the action probabilities.

The bound is sharp.
Consider two actions and three possible
states $s_1,s_2,s_0$ with posterior
probabilities $p_1,p_2,1-p_1-p_2$,
respectively.
For $i=1,2$, set
$c_g(s_i,a_i^\star)=0$ and
$c_g(s_i,a_{3-i}^\star)=\Delta_i$.
Let both action costs vanish at $s_0$.
The full-state optimal cost is zero,
whereas the optimal observation-based
cost equals
$\min\{p_1\Delta_1,p_2\Delta_2\}$.
Hence equality is attainable.

We next derive the exact decision cost of
reducing the available observation.
Fix a goal $g$ and assume integrable
action costs.
Let $H'$ be a possibly randomized
coarsening of $H$, such that
\[
c_g(S,a)\perp H'\mid H,
\qquad a\in\{a_0,a_1\}.
\]
Thus, $H'$ provides no additional
information about the action costs
once $H$ is known.

For an observation variable $T$,
define its optimal achievable risk by
\begin{equation}
R_g^\star(T)
=
\mathbb E\left[
\min_{a\in\{a_0,a_1\}}
\mathbb E[c_g(S,a)\mid T]
\right].
\label{eq:app_coarsened_optimal_risk}
\end{equation}
Write
$\Delta_g(H)
=
Q_g(a_0\mid H)-Q_g(a_1\mid H)$.
The conditional-independence assumption
and the tower property imply
\begin{equation}
\Delta_g(H')
=
\mathbb E[\Delta_g(H)\mid H'].
\label{eq:app_coarsened_contrast}
\end{equation}

Under the preceding assumptions, observation
coarsening incurs the following exact
nonnegative decision loss:
\begin{equation}
\begin{aligned}
&R_g^\star(H')-R_g^\star(H)\\
&=
\frac12\Bigl(
\mathbb E|\Delta_g(H)|
-
\mathbb E\bigl|
\mathbb E[\Delta_g(H)\mid H']
\bigr|
\Bigr)
\ge 0.
\end{aligned}
\label{eq:app_exact_coarsening_loss}
\end{equation}
Moreover, let
\[
A_\pm(H')
=
\mathbb E[(\Delta_g(H))_\pm\mid H'].
\]
Then the loss admits the equivalent
nonnegative representation
\begin{equation}
R_g^\star(H')-R_g^\star(H)
=
\mathbb E\bigl[
\min\{A_+(H'),A_-(H')\}
\bigr].
\label{eq:app_coarsening_sign_conflict}
\end{equation}
It is strictly positive if and only if
\begin{equation}
\mathbb P\bigl(
A_+(H')>0,\,
A_-(H')>0
\bigr)>0.
\label{eq:app_coarsening_strictness}
\end{equation}

To establish these identities,
recall that for any real numbers $u,v$,
\[
\min\{u,v\}
=
\frac12(u+v-|u-v|).
\]
Applying this identity to the two
conditional action costs yields
\begin{equation}
\begin{aligned}
R_g^\star(H)
=
\frac12\mathbb E\bigl[
&Q_g(a_0\mid H)
+Q_g(a_1\mid H)\\
&-|\Delta_g(H)|
\bigr].
\end{aligned}
\label{eq:app_coarsening_risk_identity}
\end{equation}
The corresponding expression for $H'$
has the same expected sum of action costs,
by the tower property.
Subtracting the two expressions
establishes
\eqref{eq:app_exact_coarsening_loss}.
Its nonnegativity follows from
conditional Jensen's inequality.

For the stronger characterization,
write
\[
\Delta_g(H)
=
(\Delta_g(H))_+
-
(\Delta_g(H))_-.
\]
Consequently,
\begin{equation}
\begin{aligned}
&\mathbb E[
|\Delta_g(H)|\mid H']
-
\left|
\mathbb E[\Delta_g(H)\mid H']
\right|\\
&\qquad=
A_+(H')+A_-(H')\\
&\qquad\quad-
|A_+(H')-A_-(H')|\\
&\qquad=
2\min\{A_+(H'),A_-(H')\}.
\end{aligned}
\label{eq:app_coarsening_positive_negative}
\end{equation}
Taking expectations proves
\eqref{eq:app_coarsening_sign_conflict}.
Since both $A_+$ and $A_-$
are nonnegative, their minimum has
strictly positive expectation exactly
under
\eqref{eq:app_coarsening_strictness}.

Thus, observation coarsening produces
additional optimal decision loss
precisely when it merges situations
with opposing nonzero
action preferences.
Losing information that does not
change the optimal action need not
increase decision risk.
This result is an exact binary-action
form of information monotonicity;
it does not require recovery
of the hidden state.

\section{Additional Decision-Limit Diagnostics}
\label{app:decision_limit_diagnostics}

Among history--goal pairs exhibiting
action conflict, the mean conditional
irreducible regrets are 2.4605
(Dev) and 2.6280 (Shifted dev),
with respective maxima of 10.56
and 13.00.

The computed oracle regret agrees
with the certified lower bound
to numerical precision, with
maximum discrepancy below
$1.2\times10^{-13}$.

\begin{table}[t]
\centering
\caption{Physical decision-limit certification
over 38,400 history--goal pairs per regime
(squared terminal-position cost).}
\label{tab:exp_decision_limits}
\setlength{\tabcolsep}{4pt}
\small
\begin{tabular}{lrr}
\toprule
Quantity & Dev & Shifted dev \\
\midrule
Observation ambiguity (\%) & 50.00 & 50.00 \\
Action conflict (\%) & 13.49 & 15.13 \\
Conflict given ambiguity (\%) & 26.98 & 30.27 \\
\midrule
Full-state optimal risk & 324.9373 & 394.6059 \\
Observation-optimal risk & 325.2692 & 395.0036 \\
Irreducible regret & 0.3319 & 0.3977 \\
Certified lower bound & 0.3319 & 0.3977 \\
\bottomrule
\end{tabular}
\end{table}

We next quantify the stability of the
optimal action under goal perturbations.
For a fixed history $h$ with
$d(h)\neq0$, the exact action-cost
contrast is
$\Delta_g(h)=\kappa(h)-2g^\top d(h)$.
For any goal perturbation
$u\in\mathbb R^2$,
\begin{equation}
\begin{aligned}
\Delta_{g+u}(h)
&=\Delta_g(h)-2u^\top d(h),\\
|\Delta_{g+u}(h)-\Delta_g(h)|
&\le 2\|u\|_2\|d(h)\|_2.
\end{aligned}
\label{eq:app_goal_perturbation}
\end{equation}

Consequently, when
$\Delta_g(h)\neq0$, the optimal
action remains unchanged for every
admissible goal $g+u$ satisfying
\begin{equation}
\|u\|_2<
r_g^{\mathrm{goal}}(h)
:=
\frac{|\Delta_g(h)|}
{2\|d(h)\|_2}.
\label{eq:app_goal_stability_radius}
\end{equation}

This radius is exact in the ambient
Euclidean goal space. Indeed, the
orthogonal projection of $g$ onto
the action-indifference boundary is
\begin{equation}
\begin{aligned}
g^\partial
&=g+
\frac{\Delta_g(h)}
{2\|d(h)\|_2^2}d(h),\\
\Delta_{g^\partial}(h)&=0,\\
\|g^\partial-g\|_2
&=r_g^{\mathrm{goal}}(h).
\end{aligned}
\label{eq:app_goal_boundary_projection}
\end{equation}

Thus, the stability radius equals
the Euclidean distance to the
decision boundary
$\{g':2{g'}^\top d(h)=\kappa(h)\}$.

The preceding goal-space stability-radius
formula assumes $d(h)\neq 0$.

When $d(h)=0$, the action-cost contrast
satisfies $\Delta_g(h)=\kappa(h)$ for
every goal $g$.
If $\kappa(h)\neq 0$, the unique optimal
action is independent of the goal, and
its goal-space stability radius is infinite.
If $\kappa(h)=0$, both actions are optimal
for every goal. In this case, the expression
$|\Delta_g(h)|/(2\|d(h)\|_2)$ is undefined
and must not be interpreted as a finite
stability radius, although the
optimal-action set remains constant.

When $d(h)\neq 0$ but $\Delta_g(h)=0$,
the goal lies on the action-indifference
boundary. Both actions are optimal at
this goal, but arbitrarily small
perturbations in the ambient
$\mathbb R^2$ can make either action
uniquely optimal. Thus, the
strict-preference stability radius is zero.

To probe the decision boundary,
consider the normal-direction
goal perturbation
\begin{equation}
g_t=g^\partial+
t\frac{d(h)}{\|d(h)\|_2}.
\label{eq:app_goal_boundary_probe}
\end{equation}
Its exact action-cost contrast is
\begin{equation}
\Delta_{g_t}(h)
=-2t\|d(h)\|_2.
\label{eq:app_goal_boundary_probe_contrast}
\end{equation}

For admissible probe goals, the
observation-optimal policy therefore
selects $a_1$ when $t<0$ and
$a_0$ when $t\geq0$, with the tie
at $t=0$ resolved in favor of $a_0$.
This gives an analytic prediction
for decision changes across the
boundary without reconstructing
the hidden physical state.

When $d(h)=0$, the contrast is
independent of $g$, so the optimal
action is constant across the
ambient goal space.

\section{Hilbert Exactness Diagnostics}
\label{app:hilbert_exactness_diagnostics}

We first establish the exact rank
of the goal-feature family.
Let $\mathcal G\subseteq\mathbb R^2$ be
a nonempty goal set, and define
$\phi(g)=(1,-2g_1,-2g_2)^\top$.
Then
\begin{equation}
\label{eq:app_goal_feature_rank}
\dim\operatorname{span}
\{\phi(g):g\in\mathcal G\}
=
1+\dim\operatorname{aff}(\mathcal G).
\end{equation}

To derive the rank identity,
fix any $g_0\in\mathcal G$.
For every $g\in\mathcal G$,
\begin{equation}
\label{eq:app_goal_feature_difference}
\phi(g)-\phi(g_0)
=
\begin{pmatrix}
0\\
-2(g-g_0)
\end{pmatrix}.
\end{equation}
Consequently,
$\operatorname{span}\phi(\mathcal G)$
is generated by $\phi(g_0)$ and the
difference vectors
$\{\phi(g)-\phi(g_0):g\in\mathcal G\}$.
The first coordinate of $\phi(g_0)$
equals one, whereas every difference
vector has first coordinate zero.
Their respective spans therefore
intersect only at the zero vector.
The dimensions add, giving
\begin{equation}
\label{eq:app_goal_rank_derivation}
\begin{aligned}
\dim\operatorname{span}\phi(\mathcal G)
&=1+\dim\operatorname{span}
  \{g-g_0:g\in\mathcal G\}\\
&=1+\dim\operatorname{aff}(\mathcal G).
\end{aligned}
\end{equation}

Thus, a singleton goal set has feature
rank one, a nontrivial collinear goal
set has rank two, and a goal set
containing three noncollinear points
has rank three.
These are exact feature-span ranks;
they do not by themselves establish
minimality among arbitrary nonlinear
decision representations.

We next establish a dimension lower bound
for exact bilinear representations.
Suppose $\mathcal G$ contains three
noncollinear goals $g_1,g_2,g_3$,
and the admissible history family contains
three histories $h_1,h_2,h_3$ whose
Hilbert states $z(h_i)\in\mathbb R^3$
are linearly independent.
Assume these histories belong to the
common domain on which the exact
contrast identity holds.

Consider any representation in an
$r$-dimensional real Hilbert space
with mappings $u(h)$ and $v(g)$ satisfying
\begin{equation}
\label{eq:app_bilinear_representation}
\Delta_g(h)=
\langle u(h),v(g)\rangle
\end{equation}
for every admissible history and goal.
Then $r\geq 3$.

To establish the lower bound,
define the matrices
\[
Z=
\begin{bmatrix}
z(h_1)&z(h_2)&z(h_3)
\end{bmatrix},
\qquad
\Phi=
\begin{bmatrix}
\phi(g_1)&\phi(g_2)&\phi(g_3)
\end{bmatrix}.
\]
By assumption, $Z$ is invertible.
The preceding goal-feature rank
result implies that $\Phi$ is invertible.

The exact action-cost contrast matrix is
\begin{equation}
\label{eq:app_contrast_matrix_rank}
\begin{aligned}
C
&=\bigl[\Delta_{g_j}(h_i)\bigr]_{i,j=1}^3\\
&=Z^\top\Phi,
\qquad \operatorname{rank}(C)=3.
\end{aligned}
\end{equation}
Choose an orthonormal basis of the
$r$-dimensional representation space.
Let $U,V\in\mathbb R^{3\times r}$
contain the coordinate vectors of
$u(h_i)$ and $v(g_j)$ as their rows.
Exact bilinearity gives
\begin{equation}
\label{eq:app_bilinear_rank_bound}
C=UV^\top,
\qquad
3=\operatorname{rank}(C)\leq r.
\end{equation}
Therefore, every representation
in this class requires at least
three dimensions.
The constructed state $z(h)$ and
goal feature $\phi(g)$ attain this
dimension, so the bound is tight.

This minimality result concerns
exact bilinear representations of
the numerical action-cost contrast.
It does not establish minimality
among arbitrary nonlinear encodings,
nor among representations preserving
only optimal-action labels.
If the attainable state family or
goal-feature family has lower rank,
a smaller exact bilinear
representation may suffice.

We verify the exact Hilbert
cost-contrast identity independently
of learned state estimators.

For each evaluation history $h_i$
and goal $g_j$, the physical oracle
computes the posterior-averaged
action-cost contrast over compatible
physical branches.

The Hilbert representation computes
the same contrast from the conditional
moment state $z(h_i)$ and the
goal feature $\phi(g_j)$.

Define the identity residual
\begin{equation}
\begin{aligned}
\varepsilon_{ij}^{\mathrm{id}}
&=\sum_{b\in\mathcal B(h_i)}
p_b(h_i)\delta_{g_j}^b(h_i)\\
&\quad-
\langle z(h_i),\phi(g_j)\rangle,\\
E_{\max}^{\mathrm{id}}
&=\max_{i,j}
|\varepsilon_{ij}^{\mathrm{id}}|.
\end{aligned}
\label{eq:app_hilbert_identity_residual}
\end{equation}

The evaluation contains 1,536
histories and 25 fixed goals
per regime, giving 38,400
history--goal pairs.

The maximum absolute identity
residuals are
$5.83\times10^{-13}$ on DEV
and $7.25\times10^{-13}$
on SHIFTED\_DEV.

The physical oracle and exact
Hilbert policy select identical
actions on all evaluated pairs,
including the prescribed
tie-breaking convention.

We next examine the finite-sample
rank structure of the action-cost
contrast matrix.
For the sampled histories and
goals, let
\begin{equation}
\begin{aligned}
Z_{i,:}&=z(h_i)^\top,\\
\Phi_{j,:}&=\phi(g_j)^\top,\\
C&=Z\Phi^\top.
\end{aligned}
\label{eq:app_hilbert_sampled_matrices}
\end{equation}

Let
$J=I_n-n^{-1}\mathbf1_n\mathbf1_n^\top$
center the histories.

In both evaluation regimes,
the numerical rank diagnostics return
\begin{equation}
\operatorname{rank}(JZ)
=\operatorname{rank}(\Phi)
=\operatorname{rank}(JC)=3.
\label{eq:app_hilbert_rank_certificate}
\end{equation}

The sampled rank test connects directly
to the restricted minimality result.
Consider an $r$-dimensional representation
that preserves the exact action-cost
contrast up to a history-independent,
goal-dependent offset:
\begin{equation}
\label{eq:app_offset_representation}
\Delta_g(h)
=
\langle u(h),v(g)\rangle+b(g).
\end{equation}
For the sampled histories and goals,
let $U$ and $V$ contain the corresponding
representation coordinates, and let
$b=(b(g_j))_j$.
Since $J\mathbf{1}_n=0$, we obtain
\begin{equation}
\label{eq:app_centered_rank_bound}
\begin{aligned}
C&=UV^\top+\mathbf{1}_n b^\top,\\
JC&=(JU)V^\top,\\
\operatorname{rank}(JC)&\leq r.
\end{aligned}
\end{equation}
Consequently, an exact rank-three
centered contrast matrix $JC$
implies $r\geq 3$, even when such
goal-dependent offsets are allowed.

The ranks reported above are numerical
diagnostics on the sampled histories
and goals. Certifying exact rank from
floating-point computations additionally
requires a verified singular-value gap
larger than the corresponding numerical
matrix error. Without this verification,
numerical rank alone does not constitute
a formal certificate of algebraic
minimality. The theoretical lower bound
remains conditional on the stated rank
assumptions and applies to exact bilinear
cost-contrast representations, not
arbitrary nonlinear encodings or
representations preserving only
optimal-action labels.

We next give a sufficient condition
for certifying exact rank three
from approximate matrices.
Let $A=JZ\in\mathbb R^{n\times 3}$ and
$B=\Phi\in\mathbb R^{m\times 3}$ denote
the exact centered-state and goal-feature
matrices, respectively, with $n\geq 4$.
Suppose their computed approximations
$\widetilde A$ and $\widetilde B$ satisfy
verified spectral-norm error bounds
$\|\widetilde A-A\|_2\leq\eta_A$ and
$\|\widetilde B-B\|_2\leq\eta_B$.

Then their smallest singular values obey
\begin{equation}
\label{eq:app_singular_value_certificate}
\begin{aligned}
\sigma_3(A)
&\geq\sigma_3(\widetilde A)-\eta_A,\\
\sigma_3(B)
&\geq\sigma_3(\widetilde B)-\eta_B.
\end{aligned}
\end{equation}
Consequently, if
$\sigma_3(\widetilde A)>\eta_A$ and
$\sigma_3(\widetilde B)>\eta_B$,
then
\begin{equation}
\label{eq:app_certified_full_rank}
\operatorname{rank}(JC)=3.
\end{equation}

To establish the singular-value bounds,
consider any unit vector $x\in\mathbb R^3$.
The reverse triangle inequality gives
\begin{equation}
\label{eq:app_singular_value_proof}
\begin{aligned}
\|Ax\|_2
&\geq\|\widetilde A x\|_2
     -\|(A-\widetilde A)x\|_2\\
&\geq\sigma_3(\widetilde A)-\eta_A.
\end{aligned}
\end{equation}
Taking the minimum over all unit vectors
proves the first inequality.
The second follows identically for $B$.

Under the stated strict inequalities,
both $A$ and $B$ have column rank three.
Since $JC=AB^\top$ and $B^\top$
has a right inverse, their product
also has rank three.

The error bounds must account for all
relevant sources of approximation,
including conditional-moment estimation
and numerical computation.
Without verified values for
$\eta_A$ and $\eta_B$, the conditions
above remain a certification criterion,
not an experimentally established
certificate.

The preceding three-dimensional construction
extends to weighted quadratic decision costs,
with the required representation dimension
controlled by the rank of the weight matrix.

Let the action outcomes belong to
$\mathbb R^m$ and let
$W\in\mathbb R^{m\times m}$
be a fixed symmetric
positive-semidefinite matrix
with rank $r\ge1$.
Consider the goal-dependent cost
\begin{equation}
c_g(s,a)
=
\mathbb E\left[
(Y_a-g)^\top W(Y_a-g)
\mid S=s
\right],
\label{eq:app_weighted_quadratic_cost}
\end{equation}
where $g\in\mathbb R^m$.
As in the unweighted model, assume
$Y_a\perp H\mid S$ for every action $a$,
and that the required conditional
second moments are finite.

Let
$U\in\mathbb R^{m\times r}$
have orthonormal columns spanning
the range of $W$.
Define
\begin{equation}
\begin{aligned}
m_a(h)
&=
\mathbb E[Y_a\mid H=h],\\
v_{a,W}(h)
&=
\mathbb E[Y_a^\top WY_a\mid H=h].
\end{aligned}
\label{eq:app_weighted_conditional_moments}
\end{equation}

The weighted quadratic decision problem
also admits an exact bilinear representation.
Define the conditional moment differences
\begin{equation}
\begin{aligned}
\kappa_W(h)
&=
v_{0,W}(h)-v_{1,W}(h),\\
d_W(h)
&=
U^\top W[m_0(h)-m_1(h)].
\end{aligned}
\label{eq:app_weighted_state_components}
\end{equation}
Then the $(r+1)$-dimensional
state and goal feature
\begin{equation}
\begin{aligned}
z_W(h)
&=
(\kappa_W(h),d_W(h)),\\
\phi_W(g)
&=
(1,-2U^\top g)
\end{aligned}
\label{eq:app_weighted_hilbert_features}
\end{equation}
satisfy the exact contrast identity
\begin{equation}
\Delta_g(h)
=
\langle z_W(h),\phi_W(g)\rangle.
\label{eq:app_weighted_exact_contrast}
\end{equation}
Thus, the optimal
observation-conditioned binary action
is determined without reconstructing
the hidden physical state.

Expanding the weighted quadratic cost
and taking conditional expectations gives
\begin{equation}
\begin{aligned}
Q_g(a\mid h)
&=
v_{a,W}(h)
-
2g^\top Wm_a(h)\\
&\quad+
g^\top Wg.
\end{aligned}
\label{eq:app_weighted_cost_expansion}
\end{equation}
Subtracting the costs of $a_0$
and $a_1$ cancels the
goal-only quadratic term:
\begin{equation}
\begin{aligned}
\Delta_g(h)
&=
\kappa_W(h)\\
&\quad-
2g^\top W[m_0(h)-m_1(h)].
\end{aligned}
\label{eq:app_weighted_contrast_expansion}
\end{equation}

Since the vector
$W[m_0(h)-m_1(h)]$
belongs to the range of $W$,
and $UU^\top$ is the orthogonal
projector onto that range,
\begin{equation}
\begin{aligned}
&g^\top W[m_0(h)-m_1(h)]\\
&\quad=
(U^\top g)^\top
U^\top W[m_0(h)-m_1(h)]\\
&\quad=
(U^\top g)^\top d_W(h).
\end{aligned}
\label{eq:app_weighted_projection_identity}
\end{equation}
Substitution into
\eqref{eq:app_weighted_contrast_expansion}
proves
\eqref{eq:app_weighted_exact_contrast}.

The weighted construction admits a matching
dimension lower bound for the unrestricted
family of affine contrast functions
\[
f_{\kappa,d}(g)
=
\kappa-2d^\top U^\top g,
\qquad
(\kappa,d)\in\mathbb R^{r+1}.
\]
Any bilinear representation that
exactly realizes every function
in this family for all
$g\in\mathbb R^m$
requires at least $r+1$
representation coordinates.

To establish this lower bound,
observe that the projected goal map
$g\mapsto U^\top g$
is surjective onto $\mathbb R^r$.
Consequently, the functions
\[
1,\quad
(U^\top g)_1,\ldots,(U^\top g)_r
\]
are linearly independent
over the full goal domain.

Choose $r+1$ goals whose projected
coordinates are
$0,e_1,\ldots,e_r$.
Evaluating these independent
functions at the selected goals
gives a matrix of rank $r+1$.

Any exact bilinear representation
with $p$ coordinates factors this
evaluation matrix through
$\mathbb R^p$, so its rank
cannot exceed $p$.
Therefore $p\ge r+1$.
The construction in
\eqref{eq:app_weighted_hilbert_features}
attains this dimension.

If the projected goals
$U^\top\mathcal G$
have affine dimension $q<r$,
the span of their weighted
goal features has dimension
\begin{equation}
\dim\operatorname{span}
\{\phi_W(g):g\in\mathcal G\}
=
1+q.
\label{eq:app_weighted_goal_family_rank}
\end{equation}
Thus, the required
contrast-preserving dimension
also depends on the
geometry of the evaluated
goal family.

The result assumes a fixed
positive-semidefinite weighting
matrix and finite conditional
second moments.
The stated lower bound concerns
exact bilinear representation
of the unrestricted affine
contrast class.
It does not establish
minimality among arbitrary
nonlinear decision-only encoders.

When $W=I_2$, the construction
reduces to the three-dimensional
Hilbert decision state
used in the main paper.
The result concerns
observation-conditioned action
comparisons and does not by itself
establish sufficiency for
multi-step control under
arbitrary future dynamics.

We next examine the boundary of exactness
for losses beyond quadratic goal costs.
The three-dimensional construction relies on
the second-order structure of the quadratic goal cost.
It does not, in general, preserve action comparisons
for higher-order losses. The following counterexample
shows that this limitation cannot be removed without
additional distributional information.

\refstepcounter{proposition}
\label{prop:app_quartic_insufficiency}
Consider the binary-action problem with the quartic
goal cost
\begin{equation}
\label{eq:app_quartic_cost}
c_g(s,a)
=
\mathbb E[
\|Y_a-g\|^4
\mid S=s
].
\end{equation}
There exist two observation histories $h_A,h_B$
with identical three-dimensional quadratic Hilbert
decision states,
\begin{equation}
\label{eq:app_quartic_same_state}
z(h_A)=z(h_B),
\end{equation}
but opposite unique observation-optimal actions
under the quartic cost at the same goal.
Consequently, the quadratic Hilbert decision state
is not universally sufficient for quartic goals.

To verify this claim, let $g=(0,0)$ and define
$Y_a=(X_a,0)\in\mathbb R^2$.
At history $h_A$, let
\begin{equation}
\label{eq:app_quartic_first_distribution}
\mathbb P(X_0=-1\mid h_A)
=
\mathbb P(X_0=1\mid h_A)
=
\frac12.
\end{equation}
At history $h_B$, instead let
\begin{equation}
\label{eq:app_quartic_second_distribution}
\begin{aligned}
\mathbb P(X_0=0\mid h_B)&=\frac23,\\
\mathbb P(X_0=\pm\sqrt{3}\mid h_B)&=\frac16
\quad\text{for each sign}.
\end{aligned}
\end{equation}
For action $a_1$, use the same conditional
distribution at both histories:
\begin{equation}
\label{eq:app_quartic_reference_distribution}
\begin{aligned}
\mathbb P(X_1=0\mid h)&=\frac12,\\
\mathbb P(X_1=\pm\sqrt{2}\mid h)&=\frac14
\quad\text{for each sign},
\end{aligned}
\end{equation}
where $h\in\{h_A,h_B\}$.

All three distributions have zero mean and
unit second moment. Therefore, at either history,
\begin{equation}
\label{eq:app_quartic_matching_moments}
m_0(h)=m_1(h)=(0,0),
\qquad
v_0(h)=v_1(h)=1.
\end{equation}
It follows that the quadratic Hilbert
decision state satisfies
\begin{equation}
\label{eq:app_quartic_zero_state}
z(h_A)=z(h_B)=(0,0,0).
\end{equation}

However, their fourth moments differ.
For action $a_0$,
\begin{equation}
\label{eq:app_quartic_fourth_moments_a0}
\mathbb E[X_0^4\mid h_A]=1,
\qquad
\mathbb E[X_0^4\mid h_B]=3,
\end{equation}
whereas action $a_1$ satisfies
\begin{equation}
\label{eq:app_quartic_fourth_moments_a1}
\mathbb E[X_1^4\mid h_A]
=
\mathbb E[X_1^4\mid h_B]
=
2.
\end{equation}

Hence, the quartic conditional action contrasts
at $g=(0,0)$ are
\begin{equation}
\label{eq:app_quartic_opposite_contrasts}
\begin{aligned}
\Delta_g^{(4)}(h_A)&=1-2=-1,\\
\Delta_g^{(4)}(h_B)&=3-2=1.
\end{aligned}
\end{equation}
The unique optimal action is therefore $a_0$
at $h_A$ and $a_1$ at $h_B$.

Thus, two histories with exactly the same
quadratic Hilbert decision state require
opposite optimal decisions under the quartic
goal cost.

The counterexample identifies a limitation of
second-order decision sufficiency, not a failure
of the quadratic representation theorem.
For squared goal costs, the conditional action
contrast depends only on first and second
moments, making the three-dimensional state
exact. Quartic costs can depend on higher-order
moments that this state does not retain.
Accordingly, exact sufficiency for more general
losses requires a separate characterization
of their decision-relevant conditional statistics.

Exact sufficiency for quartic goals can nonetheless
be restored by retaining the additional conditional
moments in the quartic cost expansion.

\refstepcounter{proposition}
\label{prop:app_exact_quartic_representation}
To construct an exact quartic decision state,
let $Y_{a_0},Y_{a_1}\in\mathbb R^2$
have finite conditional fourth moments.
Consider the quartic goal cost
\begin{equation}
\label{eq:app_quartic_representation_cost}
Q_g^{(4)}(a\mid h)
=
\mathbb E[
\|Y_a-g\|^4\mid H=h
].
\end{equation}

Define the conditional action-comparison
statistics
\begin{equation}
\label{eq:app_quartic_comparison_statistics}
\begin{aligned}
s_4(h)
&=
\mathbb E[
\|Y_{a_0}\|^4-\|Y_{a_1}\|^4
\mid H=h],\\
u_3(h)
&=
\mathbb E[
\|Y_{a_0}\|^2Y_{a_0}
-\|Y_{a_1}\|^2Y_{a_1}
\mid H=h],\\
C_2(h)
&=
\mathbb E[
Y_{a_0}Y_{a_0}^{\top}
-Y_{a_1}Y_{a_1}^{\top}
\mid H=h],\\
d_1(h)
&=
\mathbb E[
Y_{a_0}-Y_{a_1}\mid H=h].
\end{aligned}
\end{equation}

Let $\mathbb S^2$ denote the space of
real symmetric $2\times2$ matrices.
The product space
\begin{equation}
\label{eq:app_quartic_hilbert_space}
\mathcal H_4
=
\mathbb R
\times\mathbb R^2
\times\mathbb S^2
\times\mathbb R^2
\end{equation}
has dimension eight.

Define the state
\begin{equation}
\label{eq:app_quartic_hilbert_state}
z_4(h)
=
\bigl(
s_4(h),u_3(h),C_2(h),d_1(h)
\bigr)
\end{equation}
and the goal-dependent feature
\begin{equation}
\label{eq:app_quartic_goal_feature}
\phi_4(g)
=
\bigl(
1,-4g,\,
2\|g\|^2I_2+4gg^\top,\,
-4\|g\|^2g
\bigr).
\end{equation}

Equip $\mathcal H_4$ with the
product inner product, using the
Frobenius inner product on $\mathbb S^2$.
Then
\begin{equation}
\label{eq:app_quartic_exact_identity}
Q_g^{(4)}(a_0\mid h)
-
Q_g^{(4)}(a_1\mid h)
=
\langle z_4(h),\phi_4(g)\rangle_{\mathcal H_4}
\end{equation}
for every history $h$ and goal
$g\in\mathbb R^2$.

For arbitrary $y,g\in\mathbb R^2$,
the quartic cost expands as
\begin{equation}
\label{eq:app_quartic_algebra_expansion}
\begin{aligned}
\|y-g\|^4
={}&
\|y\|^4
-4\|y\|^2g^\top y\\
&+2\|g\|^2\|y\|^2
+4(g^\top y)^2\\
&-4\|g\|^2g^\top y
+\|g\|^4.
\end{aligned}
\end{equation}

Taking conditional expectations
for each action and subtracting
eliminates the common term
$\|g\|^4$.

The remaining terms yield
\begin{equation}
\label{eq:app_quartic_conditional_contrast}
\begin{aligned}
\Delta_g^{(4)}(h)
={}&s_4(h)-4g^\top u_3(h)\\
&+2\|g\|^2\operatorname{tr}C_2(h)\\
&+4g^\top C_2(h)g\\
&-4\|g\|^2g^\top d_1(h).
\end{aligned}
\end{equation}

Since $C_2(h)$ is symmetric,
\begin{equation}
\label{eq:app_quartic_matrix_inner_product}
\begin{aligned}
&2\|g\|^2\operatorname{tr}C_2(h)
+4g^\top C_2(h)g\\
&\qquad=
\left\langle
C_2(h),
2\|g\|^2I_2+4gg^\top
\right\rangle_F.
\end{aligned}
\end{equation}

Substituting this identity into
\eqref{eq:app_quartic_conditional_contrast}
gives exactly
\eqref{eq:app_quartic_exact_identity}.

Finally,
$\dim\mathbb S^2=3$, so
\begin{equation}
\label{eq:app_quartic_dimension}
\dim\mathcal H_4
=
1+2+3+2=8.
\end{equation}
Thus, the stated representation is
finite-dimensional and exactly
preserves the quartic conditional
action contrast.

The quadratic and quartic constructions
illustrate a common principle:
exact decision sufficiency depends on
the conditional statistics required by
the goal-dependent cost.

For quadratic goals, first- and
second-order conditional moments yield
the three-dimensional state.
For quartic goals, additional higher-order
moments yield the eight-dimensional
construction above.

The eight-dimensional representation preserves
quartic action contrasts exactly. A matching
dimension lower bound for an unrestricted
bilinear contrast class is established below.
This restricted minimality result does not
imply universal minimality for physical
systems, restricted goal domains, or
decision-only encoders. Nor does the
construction establish statistical learning
rates or guarantees for arbitrary
non-polynomial costs.

\refstepcounter{proposition}
\label{prop:app_quartic_bilinear_minimality}
For the lower bound, let
$\mathcal G\subseteq\mathbb R^2$
contain a nonempty open set.
Consider the unrestricted function family
\begin{equation}
\label{eq:app_quartic_affine_family}
\mathcal F_4
=
\left\{
g\mapsto
\langle\theta,\phi_4(g)\rangle_{\mathcal H_4}
:
\theta\in\mathcal H_4
\right\}.
\end{equation}

Any exact representation of every
$f\in\mathcal F_4$ in the form
\begin{equation}
\label{eq:app_quartic_bilinear_family}
f(g)=\langle T(f),\psi(g)\rangle,
\qquad
\psi(g)\in\mathbb R^k,
\end{equation}
where $\psi$ is shared across the
function family, must satisfy
$k\geq8$.
Thus, the constructed eight-dimensional
feature space has minimal dimension
within this restricted exact
bilinear representation class.

To prove the lower bound, write
$g=(x,y)$.
The eight scalar coordinates of
$\phi_4(g)$ are
\begin{equation}
\label{eq:app_quartic_feature_coordinates}
\begin{aligned}
f_1(g)&=1,\\
f_2(g)&=-4x,\\
f_3(g)&=-4y,\\
f_4(g)&=6x^2+2y^2,\\
f_5(g)&=8xy,\\
f_6(g)&=2x^2+6y^2,\\
f_7(g)&=-4x(x^2+y^2),\\
f_8(g)&=-4y(x^2+y^2).
\end{aligned}
\end{equation}

Suppose
$\sum_{i=1}^8c_if_i(g)=0$
throughout $\mathcal G$.
Because $\mathcal G$ contains an
open set, the resulting polynomial
must vanish identically.

Comparing cubic coefficients gives
\begin{equation}
\label{eq:app_quartic_cubic_rank}
c_7=c_8=0.
\end{equation}
Comparing the coefficient of $xy$
gives $c_5=0$.

The coefficients of $x^2$ and $y^2$
then yield
\begin{equation}
\label{eq:app_quartic_quadratic_rank}
\begin{pmatrix}
6&2\\
2&6
\end{pmatrix}
\begin{pmatrix}
c_4\\c_6
\end{pmatrix}
=
\begin{pmatrix}
0\\0
\end{pmatrix}.
\end{equation}
The matrix has determinant $32>0$,
so $c_4=c_6=0$.

Finally, the constant and linear
coefficients imply
$c_1=c_2=c_3=0$.
Thus, all eight coordinate
functions are linearly independent
on $\mathcal G$:
\begin{equation}
\label{eq:app_quartic_feature_rank}
\dim\operatorname{span}
\{f_1,\ldots,f_8\}=8.
\end{equation}

Now consider any representation
\eqref{eq:app_quartic_bilinear_family}.
Its functions must belong to the
linear span of the $k$ coordinate
functions of $\psi(g)$.
Therefore,
\begin{equation}
\label{eq:app_quartic_rank_lower_bound}
8
=
\dim\mathcal F_4
\leq k.
\end{equation}

The construction
\eqref{eq:app_quartic_goal_feature}
achieves dimension eight, proving
the restricted minimality claim.

The lower bound concerns preservation
of every numerical contrast in the
unrestricted function family
$\mathcal F_4$ over a goal domain
with nonempty interior.
It does not establish minimality
for the smaller family of contrasts
realized by a particular physical
system, for decision-only encoders
that preserve only action rankings,
or for restricted goal sets.

The binary construction extends to finitely many
actions by representing their conditional costs
relative to a common reference action. This
preserves exact decision sufficiency without
requiring reconstruction of the hidden physical state.

\refstepcounter{proposition}
\label{prop:app_finite_action_sufficiency}
Let
$\mathcal A=\{a_0,\ldots,a_{M-1}\}$,
where $M\geq 2$, and suppose
$Y_{a_j}\in\mathbb R^2$ has finite conditional
second moments for every action.
Under the squared goal cost
\begin{equation}
\label{eq:app_finite_action_cost}
Q_g(a_j\mid h)
=
\mathbb E[
\|Y_{a_j}-g\|^2
\mid H=h
],
\end{equation}
there exists a decision-sufficient
conditional representation in
$\mathbb R^{3(M-1)}$ that preserves
all pairwise action-cost differences
for every $g\in\mathbb R^2$.

For every action $a_j$, define
the conditional moments
\begin{equation}
\label{eq:app_finite_action_moments}
m_j(h)=\mathbb E[Y_{a_j}\mid H=h],
\qquad
v_j(h)=\mathbb E[
\|Y_{a_j}\|^2\mid H=h].
\end{equation}
Choose $a_0$ as a reference action.
For $j=1,\ldots,M-1$, construct
\begin{equation}
\label{eq:app_finite_action_state}
z_j(h)=
\begin{pmatrix}
v_j(h)-v_0(h)\\
m_j(h)-m_0(h)
\end{pmatrix}
\in\mathbb R^3,
\end{equation}
and set $z_0(h)=0$.

Using the common goal feature
\begin{equation}
\label{eq:app_finite_action_feature}
\phi(g)=
\begin{pmatrix}
1\\-2g
\end{pmatrix},
\end{equation}
expansion of the squared cost yields
\begin{equation}
\label{eq:app_finite_action_contrast}
Q_g(a_j\mid h)-Q_g(a_0\mid h)
=
\langle z_j(h),\phi(g)\rangle.
\end{equation}

Consequently, every pairwise cost
difference satisfies
\begin{equation}
\label{eq:app_finite_action_pairwise}
Q_g(a_i\mid h)-Q_g(a_j\mid h)
=
\langle z_i(h)-z_j(h),\phi(g)\rangle.
\end{equation}

Collecting the relative states gives
\begin{equation}
\label{eq:app_finite_action_stacked_state}
Z(h)=
\bigl(
z_1(h),\ldots,z_{M-1}(h)
\bigr)
\in\mathbb R^{3(M-1)}.
\end{equation}

Because all conditional action costs
share the same reference value
$Q_g(a_0\mid h)$, their ordering
is completely determined by $Z(h)$:
\begin{equation}
\label{eq:app_finite_action_optimality}
\arg\min_{0\leq j<M}Q_g(a_j\mid h)
=
\arg\min_{0\leq j<M}
\langle z_j(h),\phi(g)\rangle.
\end{equation}
Thus, $Z(h)$ preserves the entire
observation-conditioned set of
optimal actions.

The same construction also yields a finite-action
reliability bound.

Let $\widehat z_j(h)$ estimate $z_j(h)$,
with $\widehat z_0(h)=z_0(h)=0$.
Define the estimated relative-cost error
\begin{equation}
\label{eq:app_finite_action_error}
\epsilon_{j,g}(h)
=
\langle
\widehat z_j(h)-z_j(h),
\phi(g)
\rangle.
\end{equation}

Let $a^\star_g(h)$ minimize the exact
conditional cost, and let
$\widehat a_g(h)$ minimize the
estimated relative costs.

\refstepcounter{proposition}
\label{prop:app_finite_action_regret}
For every history $h$ and goal $g$,
\begin{equation}
\label{eq:app_finite_action_regret_bound}
\begin{aligned}
0
&\leq
Q_g(\widehat a_g(h)\mid h)
-
Q_g(a^\star_g(h)\mid h)\\
&\leq
2\max_{0\leq j<M}
|\epsilon_{j,g}(h)|\\
&\leq
2\|\phi(g)\|
\max_{1\le j<M}
\|\widehat z_j(h)-z_j(h)\|.
\end{aligned}
\end{equation}
Furthermore, if the exact optimal
action is unique with gap
\begin{equation}
\label{eq:app_finite_action_gap}
\gamma_g(h)
=
\min_{a_j\neq a^\star_g(h)}
\left[
Q_g(a_j\mid h)
-
Q_g(a^\star_g(h)\mid h)
\right]>0,
\end{equation}
then the sufficient condition
\begin{equation}
\label{eq:app_finite_action_stability}
\max_{0\leq j<M}
|\epsilon_{j,g}(h)|
<
\frac{\gamma_g(h)}{2}
\end{equation}
guarantees
$\widehat a_g(h)=a^\star_g(h)$.

To establish these bounds, write
$D_j=\langle z_j(h),\phi(g)\rangle$
and
$\widehat D_j=D_j+\epsilon_{j,g}(h)$.
Let $j^\star$ and $\widehat j$
index the exact and estimated
optimal actions, respectively.

Because $\widehat j$ minimizes
$\widehat D_j$,
\begin{equation}
\label{eq:app_finite_action_regret_proof}
\begin{aligned}
D_{\widehat j}-D_{j^\star}
&=
-\epsilon_{\widehat j,g}
+
(\widehat D_{\widehat j}
-\widehat D_{j^\star})
+
\epsilon_{j^\star,g}\\
&\leq
|\epsilon_{\widehat j,g}|
+
|\epsilon_{j^\star,g}|\\
&\leq
2\max_j|\epsilon_{j,g}|.
\end{aligned}
\end{equation}
The first inequality follows from
$\widehat D_{\widehat j}
\leq\widehat D_{j^\star}$.
The final bound follows from
Cauchy--Schwarz.

If the exact optimum is unique,
then every $j\neq j^\star$ satisfies
\begin{equation}
\label{eq:app_finite_action_margin_proof}
\begin{aligned}
\widehat D_j-\widehat D_{j^\star}
&=
D_j-D_{j^\star}
+\epsilon_{j,g}
-\epsilon_{j^\star,g}\\
&\geq
\gamma_g(h)
-
2\max_k|\epsilon_{k,g}(h)|.
\end{aligned}
\end{equation}
Condition
\eqref{eq:app_finite_action_stability}
makes this difference strictly positive,
so the estimated decision selects
the exact optimal action.

This extension establishes an explicit
finite-dimensional sufficient representation
for finitely many observation-conditioned
actions under squared goal costs.
The dimension $3(M-1)$ is a constructive
upper bound, not a universal minimality claim.
The reliability bound is deterministic
conditional on the estimation errors;
it does not assert a finite-sample
convergence rate.
The construction also does not
automatically establish sufficiency
for unrestricted multi-step control.

\section{Additional Learning Results}
\label{app:additional_learning_results}

This section establishes additional
learning and action-reliability bounds
and reports goal-averaged comparisons
of conditional-moment estimators
under the development and
shifted-development regimes.

Fix a training dataset $\mathcal D_n$
and its learned decision policy
$\widehat\pi_n$.
For a goal $g$ and observation history $h$,
write
$\Delta=\Delta_g(h)$ and
$\widehat\Delta=\widehat\Delta_g(h)$.
Let
$\epsilon_g(h)=\widehat\Delta-\Delta$,
consistent with
$\epsilon_g(h)=\langle e_n(h),\phi(g)\rangle$.
Both the optimal observation-based policy
$\pi^\star$ and the learned policy
$\widehat\pi_n$ select $a_0$ when
their respective contrasts are
nonpositive, breaking ties in favor of $a_0$.

Define the conditional excess regret
\[
r_{\mathrm{exc}}(h,g)
=
Q_g(\widehat\pi_n(h,g)\mid h)
-\min_{a\in\mathcal A}Q_g(a\mid h).
\]
Then
\begin{equation}
\label{eq:app_exact_excess_regret}
r_{\mathrm{exc}}(h,g)
=
|\Delta_g(h)|
\mathbf 1\{
\widehat\pi_n(h,g)\neq\pi^\star(h,g)
\}.
\end{equation}
Moreover,
\begin{equation}
\label{eq:app_excess_regret_bound}
\begin{aligned}
0\leq
R_{\mathrm{exc}}(g\mid\mathcal D_n)
&\leq\mathbb E_H|\epsilon_g(H)|\\
&\leq
\sqrt{\mathbb E_H[\epsilon_g(H)^2]}.
\end{aligned}
\end{equation}

If the learned and observation-optimal
actions agree, their conditional
cost difference is zero.
If they disagree and $\Delta\neq 0$,
the learned policy selects the
suboptimal action, whose conditional
cost exceeds the minimum by
exactly $|\Delta|$.
When $\Delta=0$, both actions have
equal conditional cost, so the
identity remains valid regardless
of the learned action.

For $\Delta\neq 0$, an action mismatch
requires the estimated contrast
to cross the decision threshold.
Consequently,
\[
\widehat\pi_n(h,g)\neq\pi^\star(h,g)
\quad\Longrightarrow\quad
|\epsilon_g(h)|\geq|\Delta_g(h)|.
\]
Multiplying this implication by
the mismatch indicator yields
$r_{\mathrm{exc}}(h,g)\leq
|\epsilon_g(h)|$.
Taking expectation over evaluation
histories proves the first upper bound.
The second follows from
Cauchy--Schwarz under a finite
second-moment assumption.

We next characterize action reliability
through the observation-conditioned
decision margin. Fix a goal $g$
and condition on the training dataset
$\mathcal D_n$.
Define the observation-conditioned
decision margin
$M_g(h)=|\Delta_g(h)|$ and the
projected estimation error
$\epsilon_g(h)=\widehat\Delta_g(h)-\Delta_g(h)$.

For any history satisfying $M_g(h)>0$,
the sufficient condition
\begin{equation}
\label{eq:app_margin_preservation}
|\epsilon_g(h)|<M_g(h)
\end{equation}
guarantees
$\widehat\pi_n(h,g)=\pi^\star(h,g)$.

Furthermore, for every $t>0$,
\begin{equation}
\label{eq:app_action_error_probability}
\begin{aligned}
&\mathbb P_H(
\widehat\pi_n(H,g)\neq\pi^\star(H,g))\\
&\quad\leq
\mathbb P_H(M_g(H)\leq t)\\
&\qquad+
\frac{\mathbb E_H[\epsilon_g(H)^2]}{t^2}.
\end{aligned}
\end{equation}

If $|\epsilon_g(h)|<|\Delta_g(h)|$
and $\Delta_g(h)\neq 0$, then
$\widehat\Delta_g(h)$ and
$\Delta_g(h)$ have the same strict sign.
The learned and observation-optimal
policies therefore select the same action.

Conversely, an action mismatch implies
either $M_g(h)\leq t$ or
$|\epsilon_g(h)|\geq t$.
Taking probabilities and applying the
union bound gives
\begin{equation}
\label{eq:app_action_error_union}
\begin{aligned}
\mathbb P_H(\widehat\pi_n\neq\pi^\star)
&\leq\mathbb P_H(M_g\leq t)\\
&\quad+\mathbb P_H(|\epsilon_g|\geq t).
\end{aligned}
\end{equation}
Markov's inequality applied to
$\epsilon_g(H)^2$ gives
\[
\mathbb P_H(|\epsilon_g|\geq t)
\leq
\frac{\mathbb E_H[\epsilon_g(H)^2]}{t^2},
\]
which establishes the claim.

A sharper excess-regret bound follows
under a condition controlling the
probability mass near the decision boundary.
Fix a goal $g$ and condition on the
training dataset $\mathcal D_n$.
Suppose that there exist constants
$C_g>0$, $\alpha\geq 0$, and $t_0>0$
such that
\begin{equation}
\label{eq:app_margin_mass}
\mathbb P_H(0<M_g(H)\leq t)
\leq C_g t^\alpha
\end{equation}
for every $0<t\leq t_0$.

Assume additionally that the projected
estimation error satisfies
\[
\|\epsilon_g\|_{L^\infty(P_H)}
\leq \eta\leq t_0.
\]
Then the goal-conditioned learning
excess regret obeys
\begin{equation}
\label{eq:app_margin_sensitive_regret}
R_{\mathrm{exc}}(g\mid\mathcal D_n)
\leq C_g\eta^{1+\alpha}.
\end{equation}

For any history with $M_g(h)>0$,
an action mismatch requires
\[
M_g(h)\leq|\epsilon_g(h)|
\leq\eta.
\]
Using the exact binary excess-regret
identity established above gives
\begin{equation}
\label{eq:app_margin_regret_proof}
\begin{aligned}
&R_{\mathrm{exc}}(g\mid\mathcal D_n)\\
&\leq
\mathbb E_H\!\left[
M_g(H)
\mathbf 1_{\{0<M_g(H)\leq\eta\}}
\right]\\
&\leq
\eta\,\mathbb P_H(
0<M_g(H)\leq\eta)\\
&\leq C_g\eta^{1+\alpha}.
\end{aligned}
\end{equation}
Histories with $M_g(H)=0$ contribute
zero excess regret regardless of
the selected action.

The improved bound is conditional on
the margin-mass assumption and a
uniform projected-error bound.
It does not establish a sample-size
convergence rate without an independent
bound on the learned estimator.

We next extend the reliability guarantee
uniformly over a bounded family of goals.
Fix the training dataset $\mathcal D_n$.
Let $\mathcal G\subset\mathbb R^2$
be a nonempty bounded goal family,
and define
\begin{equation}
\label{eq:app_goal_feature_bound}
B_{\mathcal G}
=
\sup_{g\in\mathcal G}
\|\phi(g)\|_2
<\infty.
\end{equation}
For each observation history $h$,
let
$e_n(h)=\widehat z_n(h)-z(h)$.

Then the learned action-cost contrasts
satisfy the simultaneous bound
\begin{equation}
\label{eq:app_uniform_contrast_error}
\sup_{g\in\mathcal G}
|\widehat\Delta_g(h)-\Delta_g(h)|
\leq
B_{\mathcal G}\|e_n(h)\|_2.
\end{equation}

Consequently, if
\begin{equation}
\label{eq:app_uniform_margin_condition}
\inf_{g\in\mathcal G}
|\Delta_g(h)|
>
B_{\mathcal G}\|e_n(h)\|_2,
\end{equation}
then
$\widehat\pi_n(h,g)=\pi^\star(h,g)$
simultaneously for every
$g\in\mathcal G$.

Furthermore, the goal-uniform
learning excess regret satisfies
\begin{equation}
\label{eq:app_uniform_excess_regret}
\begin{aligned}
&\sup_{g\in\mathcal G}
R_{\mathrm{exc}}(g\mid\mathcal D_n)\\
&\quad\leq B_{\mathcal G}
\left(
\mathbb E_H\|e_n(H)\|_2^2
\right)^{1/2}.
\end{aligned}
\end{equation}

By exact Hilbert contrast factorization,
\[
\begin{aligned}
|\widehat\Delta_g(h)-\Delta_g(h)|
&=|\langle e_n(h),\phi(g)\rangle|\\
&\leq\|e_n(h)\|_2\|\phi(g)\|_2.
\end{aligned}
\]
Taking the supremum over
$g\in\mathcal G$ proves
\eqref{eq:app_uniform_contrast_error}.

Under
\eqref{eq:app_uniform_margin_condition},
every true contrast has magnitude
strictly greater than its estimation
error. Its sign is therefore preserved
for every admissible goal, proving
simultaneous action agreement.

Finally, the exact binary excess-regret
bound gives, for each goal,
\[
\begin{aligned}
R_{\mathrm{exc}}(g\mid\mathcal D_n)
&\leq\mathbb E_H|\epsilon_g(H)|\\
&\leq B_{\mathcal G}
\mathbb E_H\|e_n(H)\|_2\\
&\leq B_{\mathcal G}
\left(
\mathbb E_H\|e_n(H)\|_2^2
\right)^{1/2}.
\end{aligned}
\]
The right-hand side is independent
of $g$, yielding the uniform bound.

For the goal feature used here,
$\phi(g)=(1,-2g_1,-2g_2)$, so
\[
B_{\mathcal G}
=
\left(
1+4\sup_{g\in\mathcal G}\|g\|_2^2
\right)^{1/2}.
\]
A uniform guarantee requires a
bounded goal family and
state-estimation error.
Simultaneous action agreement additionally
requires the strict margin separation
in \eqref{eq:app_uniform_margin_condition}.

We next examine the finite-sample
performance of the estimators under
the two registered evaluation regimes.

\begin{table}[H]
\centering
\caption{Goal-averaged excess regret on 25 registered goals
at $n=6912$ (lower is better).}
\label{tab:exp43_shrinkage}
\setlength{\tabcolsep}{5pt}
\begin{tabular}{lcc}
\toprule
Estimator & DEV & SHIFTED\_DEV\\
\midrule
Mean-only ($\lambda=0$)
    & 0.09209 & 0.08186\\
Full moment ($\lambda=1$)
    & 0.12800 & 0.16057\\
CV shrinkage
    & 0.08014 & 0.09518\\
\bottomrule
\end{tabular}
\end{table}

Table~\ref{tab:exp43_shrinkage}
shows that estimator performance
depends on the evaluation regime.
On DEV, cross-validated shrinkage
achieves the lowest goal-averaged
excess regret (0.08014), compared
with 0.09209 for the mean-only
estimator and 0.12800 for the
full-moment estimator.

On SHIFTED\_DEV, however, the
mean-only estimator achieves the
lowest regret (0.08186), followed
by cross-validated shrinkage
(0.09518) and the full-moment
estimator (0.16057).

These results establish neither
uniform dominance of moment-based
estimation nor optimality of the
cross-validated shrinkage coefficient
under distribution shift.
They are consistent with a
finite-sample estimation trade-off,
but the aggregate comparison alone
does not identify its precise cause.

The theoretical exactness of the
Hilbert decision state is therefore
distinct from the finite-sample
reliability of its learned estimate.

We select the shrinkage coefficient
using leave-one-simulator-seed-out
cross-validation on the training data.

For each held-out seed $s$, conditional
moment estimators are fitted using
training histories from the remaining
seeds. Let $\mathcal I_s$ index the
held-out histories and let
$g_1,\ldots,g_K$ denote the registered
training goals.

For each held-out history $h_i$,
define
\begin{equation}
\begin{aligned}
A_{sij}
&=\widehat\kappa_{m,-s}(h_i)
-2g_j^\top\widehat d_{-s}(h_i)\\
&\quad-\Delta_{g_j}(h_i),\\
B_{si}
&=\widehat\tau_{-s}(h_i).
\end{aligned}
\label{eq:app_cv_residual_terms}
\end{equation}
Here, the subscript $-s$ denotes
estimation without seed $s$.
The reference contrast
$\Delta_{g_j}(h_i)$ is computed from
the training simulator.

The held-out projected-contrast
error is
\begin{equation}
\begin{aligned}
\widehat{\mathcal L}_{\mathrm{CV}}
(\lambda)
&=\sum_s\sum_{i\in\mathcal I_s}
\sum_{j=1}^{K}\\
&\quad
(A_{sij}+\lambda B_{si})^2.
\end{aligned}
\label{eq:app_cv_shrinkage_objective}
\end{equation}
A positive normalization factor
does not affect its minimizer.

Define
\begin{equation}
\begin{aligned}
a_{\mathrm{CV}}
&=\sum_s\sum_{i\in\mathcal I_s}
\sum_{j=1}^{K}
A_{sij}B_{si},\\
b_{\mathrm{CV}}
&=K\sum_s\sum_{i\in\mathcal I_s}
B_{si}^2.
\end{aligned}
\label{eq:app_cv_quadratic_terms}
\end{equation}

When $b_{\mathrm{CV}}>0$,
the minimizer over $\lambda\in[0,1]$
has the closed form
\begin{equation}
\widehat\lambda_{\mathrm{CV}}
=
\Pi_{[0,1]}
\left(
-\frac{a_{\mathrm{CV}}}
{b_{\mathrm{CV}}}
\right),
\label{eq:app_cv_shrinkage_solution}
\end{equation}
where $\Pi_{[0,1]}$ denotes
projection onto $[0,1]$.

If $b_{\mathrm{CV}}=0$, the objective
is independent of $\lambda$ and
does not identify a unique
shrinkage coefficient.

This selection procedure uses
only training histories and
registered training goals.
Development and shifted-development
samples are excluded from
hyperparameter selection.

The preceding uniform error bound
holds for general bounded goal families.
For the three-dimensional Hilbert
decision state, the affine dependence
on the goal permits an exact
characterization.

Fix a history $h$ and decompose
the state-estimation error as
\[
e_n(h)
=
(e_\kappa(h),e_d(h)),
\qquad
e_d(h)\in\mathbb R^2.
\]
Since
$\phi(g)=(1,-2g_1,-2g_2)^\top$,
the projected contrast error is
\begin{equation}
\epsilon_g(h)
=
e_\kappa(h)-2e_d(h)^\top g.
\label{eq:app_affine_goal_error}
\end{equation}

For a nonempty compact goal set
$\mathcal G\subset\mathbb R^2$,
define its support function by
\begin{equation}
\sigma_{\mathcal G}(u)
=
\sup_{g\in\mathcal G}u^\top g.
\label{eq:app_goal_support_function}
\end{equation}

The support function gives an exact
expression for the maximum projected
error over $\mathcal G$:
\begin{equation}
\begin{aligned}
U_{\mathcal G}(h)
&:=
\sup_{g\in\mathcal G}
|\epsilon_g(h)|\\
&=
\max\{U_+(h),U_-(h)\},
\end{aligned}
\label{eq:app_exact_goal_error}
\end{equation}
where
\begin{equation}
\begin{aligned}
U_+(h)
&=
e_\kappa(h)
+
2\sigma_{\mathcal G}(-e_d(h)),\\
U_-(h)
&=
-e_\kappa(h)
+
2\sigma_{\mathcal G}(e_d(h)).
\end{aligned}
\label{eq:app_goal_error_extrema}
\end{equation}

To establish the identity, note that
for any real-valued function $f$,
\[
\sup_g|f(g)|
=
\max\{\sup_g f(g),\sup_g[-f(g)]\}.
\]
Applying this identity to
\eqref{eq:app_affine_goal_error}
gives
\begin{equation}
\begin{aligned}
\sup_{g\in\mathcal G}\epsilon_g(h)
&=
e_\kappa(h)
+
2\sigma_{\mathcal G}(-e_d(h)),\\
\sup_{g\in\mathcal G}[-\epsilon_g(h)]
&=
-e_\kappa(h)
+
2\sigma_{\mathcal G}(e_d(h)).
\end{aligned}
\label{eq:app_goal_error_proof}
\end{equation}
Taking their maximum proves
\eqref{eq:app_exact_goal_error}.

In particular, consider the Euclidean
goal family
\[
\mathcal G
=
\{g:\|g-g_0\|_2\le\rho\},
\qquad \rho\ge0.
\]
Its support function is
\[
\sigma_{\mathcal G}(u)
=
u^\top g_0+\rho\|u\|_2.
\]
Substitution gives the exact formula
\begin{equation}
\begin{aligned}
U_{\mathcal G}(h)
&=
\left|
e_\kappa(h)
-
2e_d(h)^\top g_0
\right|\\
&\quad+
2\rho\|e_d(h)\|_2.
\end{aligned}
\label{eq:app_goal_ball_exact_error}
\end{equation}
The bound is attained because
$\mathcal G$ is compact.

Uniform action preservation follows
whenever the true action contrasts satisfy
\begin{equation}
\inf_{g\in\mathcal G}
|\Delta_g(h)|
>
U_{\mathcal G}(h),
\label{eq:app_exact_uniform_margin}
\end{equation}
then
\begin{equation}
\widehat\pi_n(h,g)
=
\pi^\star(h,g),
\qquad
\forall g\in\mathcal G.
\label{eq:app_exact_goal_action_preservation}
\end{equation}

Indeed, for every $g\in\mathcal G$,
\[
|\widehat\Delta_g(h)-\Delta_g(h)|
\le U_{\mathcal G}(h)
<
|\Delta_g(h)|.
\]
Hence the estimated and true
contrasts have the same nonzero sign.
The binary action rule therefore
selects the same action for every
goal in $\mathcal G$.

The support-function identity is exact
for the specified goal set and
state-estimation error.
It does not imply that the error
is known during deployment.
An operational decision certificate
requires an independently justified
bound on the estimation error.
Moreover, if the goal family crosses
a zero-margin decision boundary,
the strict uniform condition may fail
even when individual goals remain
reliably decidable.

A uniform action-preservation condition
may be inconclusive when the goal set
intersects a zero-contrast decision
boundary.
Nevertheless, the affine contrast
structure yields a distributional
reliability guarantee.

Fix a history $h$ and suppose
the goal $G$ is uniformly distributed
over the Euclidean ball
\[
\mathcal G
=
\{g\in\mathbb R^2:
\|g-g_0\|_2\le\rho\},
\qquad \rho>0.
\]
Assume $d(h)\neq0$, where
\[
\Delta_g(h)
=
\kappa(h)-2d(h)^\top g.
\]
Let
$\eta(h)=U_{\mathcal G}(h)$
denote the exact uniform contrast-error
bound from
\eqref{eq:app_exact_goal_error}.

Under these conditions, the
goal-averaged action-disagreement
probability satisfies
\begin{equation}
\begin{aligned}
&\mathbb P_G\bigl(
\widehat\pi_n(h,G)
\neq\pi^\star(h,G)
\bigr)\\
&\quad\le
\min\left\{
1,\,
\frac{2\eta(h)}
{\pi\rho\|d(h)\|_2}
\right\}.
\end{aligned}
\label{eq:app_goal_volume_disagreement}
\end{equation}
The goal-averaged conditional
excess regret also satisfies
\begin{equation}
\begin{aligned}
&\mathbb E_G\bigl[
|\Delta_G(h)|
\mathbf 1\{
\widehat\pi_n(h,G)
\neq\pi^\star(h,G)
\}
\bigr]\\
&\quad\le
\eta(h)
\min\left\{
1,\,
\frac{2\eta(h)}
{\pi\rho\|d(h)\|_2}
\right\}.
\end{aligned}
\label{eq:app_goal_volume_regret}
\end{equation}

To establish these bounds, note that
a disagreement between the estimated
and optimal binary actions implies
\[
|\Delta_g(h)|
\le
|\widehat\Delta_g(h)-\Delta_g(h)|
\le\eta(h).
\]
Therefore, disagreement can occur
only in the goal region
\begin{equation}
\mathcal S_\eta(h)
=
\left\{
g\in\mathcal G:
|\kappa(h)-2d(h)^\top g|
\le\eta(h)
\right\}.
\label{eq:app_goal_disagreement_strip}
\end{equation}
Because $d(h)\neq0$,
this region is the intersection
of the goal ball with a strip
perpendicular to $d(h)$.

The strip has width
\[
w(h)
=
\frac{\eta(h)}{\|d(h)\|_2}.
\]
Every cross-section of the ball
parallel to the decision boundary
has length at most $2\rho$.
Integrating these cross-sections
over the strip width gives
\begin{equation}
\operatorname{Area}
(\mathcal S_\eta(h))
\le
2\rho\,w(h)
=
\frac{2\rho\eta(h)}
{\|d(h)\|_2}.
\label{eq:app_goal_strip_area}
\end{equation}
Since the ball has area
$\pi\rho^2$, uniform sampling
implies
\eqref{eq:app_goal_volume_disagreement},
together with the trivial
probability bound of one.

On every disagreement,
$|\Delta_g(h)|\le\eta(h)$.
Multiplying the disagreement
probability bound by $\eta(h)$
proves
\eqref{eq:app_goal_volume_regret}.

For fixed $\rho>0$ and
$\|d(h)\|_2>0$,
the disagreement bound is linear
in $\eta(h)$, while the
excess-regret bound is quadratic
for sufficiently small errors.

This conclusion does not require
a strictly positive margin
uniformly over the goal ball.
It instead exploits the geometry
of the affine decision boundary
and the specified uniform goal law.

The result is conditional on the
fixed history and goal distribution.
It does not extend unchanged to
arbitrary goal distributions,
and the displayed bounds are
not applicable when $d(h)=0$.
No estimator convergence rate
is asserted without a separate
bound on $\eta(h)$.

The preceding results use either
uniform contrast-error control
or a specified goal distribution.
We now derive a complementary
regret bound directly from
mean-squared contrast error.

Fix a goal $g$ and condition
on the fitted estimator.
Define
\begin{equation}
V_g
=
\mathbb E_H[
\epsilon_g(H)^2
],
\qquad
\epsilon_g(H)
=
\widehat\Delta_g(H)-\Delta_g(H).
\label{eq:app_squared_contrast_error}
\end{equation}
Assume that there exist constants
$C_g>0$, $\alpha\ge0$, and $t_0>0$
such that
\begin{equation}
\mathbb P_H\bigl(
0<|\Delta_g(H)|\le t
\bigr)
\le
C_g t^\alpha,
\quad 0<t\le t_0.
\label{eq:app_squared_error_margin_condition}
\end{equation}

Under the stated margin condition,
for every $0<t\le t_0$,
the excess regret satisfies
\begin{equation}
R_{\mathrm{exc}}(g)
\le
C_g t^{1+\alpha}
+
\frac{V_g}{t}.
\label{eq:app_squared_error_regret_tradeoff}
\end{equation}
Consequently,
\begin{equation}
R_{\mathrm{exc}}(g)
\le
\inf_{0<t\le t_0}
\left\{
C_g t^{1+\alpha}
+
\frac{V_g}{t}
\right\}.
\label{eq:app_squared_error_regret_infimum}
\end{equation}

If $V_g>0$ and
\begin{equation}
t_\star
=
\left(
\frac{V_g}
{(1+\alpha)C_g}
\right)^{\frac{1}{2+\alpha}}
\le t_0,
\label{eq:app_squared_error_optimal_threshold}
\end{equation}
then
\begin{equation}
R_{\mathrm{exc}}(g)
\le
c_\alpha
C_g^{\frac{1}{2+\alpha}}
V_g^{\frac{1+\alpha}{2+\alpha}},
\label{eq:app_squared_error_fast_regret}
\end{equation}
where
\begin{equation}
c_\alpha
=
(2+\alpha)
(1+\alpha)^{
-\frac{1+\alpha}{2+\alpha}
}.
\label{eq:app_squared_error_margin_constant}
\end{equation}
When $V_g=0$, the excess regret
is zero.

To establish the bound, let
$M(H)=|\Delta_g(H)|$
and denote the action-disagreement
event by
\[
A=
\{
\widehat\pi_n(H,g)
\neq
\pi^\star(H,g)
\}.
\]
The exact binary regret identity gives
\begin{equation}
R_{\mathrm{exc}}(g)
=
\mathbb E_H[
M(H)\mathbf 1_A
].
\label{eq:app_squared_error_regret_identity}
\end{equation}
Split the expectation according
to whether $M(H)\le t$ or $M(H)>t$.

For the small-margin contribution,
\begin{equation}
\begin{aligned}
&\mathbb E_H[
M(H)\mathbf 1_A
\mathbf 1_{\{M(H)\le t\}}
]\\
&\qquad\le
t\,\mathbb P_H(0<M(H)\le t)\\
&\qquad\le
C_g t^{1+\alpha}.
\end{aligned}
\label{eq:app_squared_error_small_margin}
\end{equation}

For the large-margin contribution,
action disagreement implies
$|\epsilon_g(H)|\ge M(H)$.
Whenever $M(H)>t$,
\[
M(H)\mathbf 1_A
\le
\frac{\epsilon_g(H)^2}{M(H)}
\le
\frac{\epsilon_g(H)^2}{t}.
\]
Therefore,
\begin{equation}
\begin{aligned}
&\mathbb E_H[
M(H)\mathbf 1_A
\mathbf 1_{\{M(H)>t\}}
]\\
&\qquad\le
\frac{V_g}{t}.
\end{aligned}
\label{eq:app_squared_error_large_margin}
\end{equation}

Adding the two contributions proves
\eqref{eq:app_squared_error_regret_tradeoff}.
Minimizing its right-hand side
over positive $t$ gives $t_\star$.
When $t_\star\le t_0$,
substitution and simplification
yield
\eqref{eq:app_squared_error_fast_regret}.
If $V_g=0$, the contrast error
vanishes almost surely, so the
learned rule has zero excess regret.

Because
$\epsilon_g(H)
=
\langle e_n(H),\phi(g)\rangle$,
Cauchy--Schwarz gives
\begin{equation}
V_g
\le
\|\phi(g)\|^2
\mathbb E_H\|e_n(H)\|^2.
\label{eq:app_squared_state_error_projection}
\end{equation}
Thus, a justified mean-squared
Hilbert state-error bound,
together with the margin condition,
controls the learning-induced
decision regret.

The result holds without assuming
independence between estimation error
and decision margin.
The improved exponent depends on
the stated margin condition,
and the optimized bound requires
$t_\star\le t_0$.
It does not establish a sample-size
convergence rate unless a separate
rate for $V_g$ is proved.

The preceding regret bounds are conditional on
the accuracy of the learned decision state.
A statistical convergence rate requires
additional assumptions on how that state
is estimated. We give an explicit benchmark
for an idealized conditional-sampling model.

\refstepcounter{proposition}
\label{prop:app_conditional_concentration}
For a finite-sample concentration benchmark,
fix a history $h$ and suppose that, for each
action $a\in\{a_0,a_1\}$, we observe $n$
conditionally independent and identically
distributed samples
$Y_{a,1},\ldots,Y_{a,n}$ from the law of
$Y_a\mid H=h$.

Assume $\|Y_a\|\leq B$ almost surely,
and let the goal family satisfy
$\mathcal G\subseteq\{g:\|g\|\leq G\}$.

Define the empirical conditional moments
\begin{equation}
\label{eq:app_empirical_conditional_moments}
\widehat m_a(h)
=
\frac1n\sum_{i=1}^nY_{a,i},
\qquad
\widehat v_a(h)
=
\frac1n\sum_{i=1}^n\|Y_{a,i}\|^2.
\end{equation}
Construct $\widehat z_n(h)$ from these
moments using the exact quadratic
Hilbert-state formula.

Then, for every $\delta\in(0,1)$,
with conditional probability at least
$1-\delta$,
\begin{equation}
\label{eq:app_uniform_finite_sample_contrast}
\sup_{g\in\mathcal G}
|\widehat\Delta_g(h)-\Delta_g(h)|
\leq \eta_{n,\delta},
\end{equation}
where
\begin{equation}
\label{eq:app_concentration_radius}
\eta_{n,\delta}
=
(2B^2+8\sqrt{2}BG)
\sqrt{\frac{\log(12/\delta)}{2n}}.
\end{equation}
To derive the concentration bound, set
\begin{equation}
\label{eq:app_concentration_t}
t=
\sqrt{\frac{\log(12/\delta)}{2n}}.
\end{equation}

Since $\|Y_a\|^2\in[0,B^2]$,
Hoeffding's inequality gives
\begin{equation}
\label{eq:app_concentration_second_moment}
\mathbb P\left(
|\widehat v_a-v_a|>B^2t
\mid H=h
\right)
\leq 2e^{-2nt^2}.
\end{equation}

Likewise, every coordinate of $Y_a$
belongs to $[-B,B]$. Hence, for
$k\in\{1,2\}$,
\begin{equation}
\label{eq:app_concentration_first_moment}
\mathbb P\left(
|\widehat m_{a,k}-m_{a,k}|>2Bt
\mid H=h
\right)
\leq 2e^{-2nt^2}.
\end{equation}

There are six scalar quantities:
two second moments and four
first-moment coordinates.
A union bound shows that all six
inequalities hold simultaneously
with probability at least
\begin{equation}
\label{eq:app_concentration_union}
1-12e^{-2nt^2}=1-\delta.
\end{equation}

On this event, the error in the
scalar contrast component satisfies
\begin{equation}
\label{eq:app_concentration_scalar_error}
|\widehat\kappa-\kappa|
\leq 2B^2t.
\end{equation}
For the vector component,
\begin{equation}
\label{eq:app_concentration_vector_error}
\|\widehat d-d\|
\leq 4\sqrt{2}Bt.
\end{equation}

Using the exact affine contrast identity,
for every $g\in\mathcal G$,
\begin{equation}
\label{eq:app_concentration_contrast_proof}
\begin{aligned}
|\widehat\Delta_g-\Delta_g|
&=
|(\widehat\kappa-\kappa)
-2g^\top(\widehat d-d)|\\
&\leq
2B^2t+8\sqrt{2}BGt.
\end{aligned}
\end{equation}
The bound is independent of the
particular goal, so it holds
simultaneously throughout $\mathcal G$.

\refstepcounter{corollary}
\label{cor:app_conditional_sampling_regret}
The concentration result also yields
a finite-sample regret guarantee under
a margin condition.
Suppose the conditional-sampling model
above is applied at each evaluated history.
For a fixed goal $g$, assume
\begin{equation}
\label{eq:app_sampling_margin_assumption}
\mathbb P_H(
0<|\Delta_g(H)|\leq t
)
\leq C_g t^\alpha
\end{equation}
for all $0<t\leq t_0$, where
$C_g>0$, $\alpha\geq0$, and $t_0>0$.

If $\eta_{n,\delta}\leq t_0$,
the expected learning-induced regret,
averaged over histories and conditional
sampling, satisfies
\begin{equation}
\label{eq:app_sampling_regret_rate}
R_{\mathrm{exc}}^{(n)}(g)
\leq
C_g\eta_{n,\delta}^{1+\alpha}
+
\delta(B+G)^2.
\end{equation}
To establish the bound, let
$\mathcal E$ be the event on which
the uniform contrast bound holds.
Conditional on any history,
$\mathbb P(\mathcal E^c\mid H)\leq\delta$.

On $\mathcal E$, an incorrect action
can be selected only when
$0<|\Delta_g(H)|\leq\eta_{n,\delta}$.
The corresponding regret is at most
$\eta_{n,\delta}$. Therefore,
\begin{equation}
\label{eq:app_sampling_good_event_regret}
\begin{aligned}
\mathbb E[
r_{\mathrm{exc}}(H,g)
\mathbf 1_{\mathcal E}
]
&\leq
\eta_{n,\delta}
\mathbb P_H(
0<|\Delta_g(H)|
\leq\eta_{n,\delta})\\
&\leq
C_g\eta_{n,\delta}^{1+\alpha}.
\end{aligned}
\end{equation}

Since $\|Y_a\|\leq B$ and $\|g\|\leq G$,
each conditional squared cost belongs to
$[0,(B+G)^2]$.
Thus, on the complementary event,
\begin{equation}
\label{eq:app_sampling_failure_event_regret}
\mathbb E[
r_{\mathrm{exc}}(H,g)
\mathbf 1_{\mathcal E^c}
]
\leq
\delta(B+G)^2.
\end{equation}

Adding the two contributions establishes
\eqref{eq:app_sampling_regret_rate}.

The concentration result establishes
a concrete statistical route from
conditional-moment estimation to
decision reliability. It uses
bounded outcomes and independent
replicated samples conditional on
each evaluated history. Such samples
are not generally available in a
partially observed trajectory.

Consequently, this result is an
idealized statistical benchmark,
not a convergence guarantee for
the learned estimators evaluated
in this paper. Establishing an
analogous rate for those estimators
requires a separate analysis of
their training data, function class,
and generalization error.

The closed-form shrinkage coefficient
follows by expanding the validation
criterion:
\begin{equation}
\label{eq:app_cv_quadratic_expansion}
\begin{aligned}
\widehat{\mathcal L}_{\mathrm{CV}}(\lambda)
&=c_{\mathrm{CV}}
+2a_{\mathrm{CV}}\lambda
+b_{\mathrm{CV}}\lambda^2,\\
c_{\mathrm{CV}}
&=\sum_{i,j}A_{ij}^2,\\
a_{\mathrm{CV}}
&=\sum_{i,j}A_{ij}B_{ij},\\
b_{\mathrm{CV}}
&=\sum_{i,j}B_{ij}^2\geq 0.
\end{aligned}
\end{equation}

When $b_{\mathrm{CV}}>0$, this is
a strictly convex quadratic with
derivative
\[
\frac{d\widehat{\mathcal L}_{\mathrm{CV}}}
{d\lambda}
=
2a_{\mathrm{CV}}
+2b_{\mathrm{CV}}\lambda.
\]
Its unconstrained minimizer is
$-a_{\mathrm{CV}}/b_{\mathrm{CV}}$.
Projection onto $[0,1]$ therefore
gives the unique constrained minimizer.

When $b_{\mathrm{CV}}=0$,
every $B_{ij}=0$, which also implies
$a_{\mathrm{CV}}=0$.
The criterion is then constant in
$\lambda$, and selecting
$\widehat\lambda_{\mathrm{CV}}=0$
is a deterministic tie-breaking
convention.

Thus the selected coefficient satisfies
\begin{equation}
\label{eq:app_cv_endpoint_guarantee}
\begin{aligned}
\widehat{\mathcal L}_{\mathrm{CV}}
(\widehat\lambda_{\mathrm{CV}})
&\leq
\widehat{\mathcal L}_{\mathrm{CV}}(0),\\
\widehat{\mathcal L}_{\mathrm{CV}}
(\widehat\lambda_{\mathrm{CV}})
&\leq
\widehat{\mathcal L}_{\mathrm{CV}}(1).
\end{aligned}
\end{equation}

This guarantee concerns the empirical
squared-contrast validation criterion,
not the evaluation regret.
Minimizing the former does not
establish minimization of the latter.

Moreover, the definition of
$A_{ij}$ uses the simulator-provided
reference contrast
$\Delta_{g_j}(h_i)$.
The selection procedure therefore
requires reference action contrasts
on the training folds.
It does not by itself establish
a label-free shrinkage selection
method for deployment when
counterfactual action costs
are unavailable.

For each evaluation configuration,
let $N$ denote the number of histories
and $K$ the number of evaluated goals.

Define the projected estimation
error and goal matrix by
\begin{equation}
\begin{aligned}
e_i
&=\widehat z(h_i)-z(h_i),\\
\epsilon_{ij}
&=\langle e_i,\phi(g_j)\rangle,\\
M_{\mathcal G}
&=\frac1K\sum_{j=1}^{K}
\phi(g_j)\phi(g_j)^\top.
\end{aligned}
\label{eq:app_empirical_error_geometry}
\end{equation}

The empirical excess regret is
\begin{equation}
\begin{aligned}
\widehat{\bar R}_{\mathrm{exc}}
&=\frac1{NK}\sum_{i=1}^{N}
\sum_{j=1}^{K}
|\Delta_{ij}|\\
&\quad\times
\mathbf1\{
\widehat\pi_{ij}\neq
\pi^\star_{ij}
\}.
\end{aligned}
\label{eq:app_empirical_excess_regret}
\end{equation}

Define the corresponding
error-based upper bounds:
\begin{equation}
\begin{aligned}
B_1
&=\frac1{NK}\sum_{i,j}
|\epsilon_{ij}|,\\
B_2
&=\sqrt{
\frac1N\sum_{i=1}^{N}
e_i^\top M_{\mathcal G}e_i
}.
\end{aligned}
\label{eq:app_empirical_regret_bounds}
\end{equation}

By the binary-action regret
identity and Cauchy--Schwarz,
\begin{equation}
0\le
\widehat{\bar R}_{\mathrm{exc}}
\le B_1\le B_2.
\label{eq:app_empirical_bound_chain}
\end{equation}

The final bound uses the exact
quadratic identity
\begin{equation}
B_2^2
=\frac1{NK}\sum_{i,j}
\epsilon_{ij}^2.
\label{eq:app_empirical_quadratic_identity}
\end{equation}

We verify these inequalities
across 168 evaluation
configurations: seven training
runs, two evaluation regimes,
four goal families, and
three estimators.
No numerical inequality
violations are observed.

To establish these finite-array bounds,
consider any fixed collection of
evaluation histories
$h_1,\ldots,h_N$ and registered goals
$g_1,\ldots,g_K$.
For each history--goal pair, define
\[
\Delta_{ij}=\Delta_{g_j}(h_i),
\qquad
\epsilon_{ij}
=\widehat\Delta_{g_j}(h_i)-\Delta_{ij}.
\]

Let
$I_{ij}=
\mathbf 1\{
\widehat\pi_n(h_i,g_j)
\neq\pi^\star(h_i,g_j)
\}$.

For every pair,
\begin{equation}
\label{eq:app_empirical_pairwise_regret}
|\Delta_{ij}|I_{ij}
\leq|\epsilon_{ij}|.
\end{equation}

To verify this inequality, first
suppose that the two actions agree.
Then $I_{ij}=0$, so the inequality
holds immediately.

If the actions disagree and
$\Delta_{ij}\neq0$, the estimated
contrast must lie on the opposite
side of zero or at the decision tie.
Consequently,
\[
|\widehat\Delta_{g_j}(h_i)-\Delta_{ij}|
\geq|\Delta_{ij}|.
\]
If $\Delta_{ij}=0$, the left-hand side
of \eqref{eq:app_empirical_pairwise_regret}
is zero regardless of the selected action.

Averaging the pairwise inequality gives
\begin{equation}
\label{eq:app_empirical_first_bound}
\begin{aligned}
\widehat{\bar R}_{\mathrm{exc}}
&=
\frac{1}{NK}
\sum_{i=1}^{N}\sum_{j=1}^{K}
|\Delta_{ij}|I_{ij}\\
&\leq
\frac{1}{NK}
\sum_{i=1}^{N}\sum_{j=1}^{K}
|\epsilon_{ij}|
=B_1.
\end{aligned}
\end{equation}

Cauchy--Schwarz gives
\begin{equation}
\label{eq:app_empirical_second_bound}
\begin{aligned}
B_1^2
&=
\left(
\frac{1}{NK}
\sum_{i,j}|\epsilon_{ij}|
\right)^2\\
&\leq
\frac{1}{NK}
\sum_{i,j}\epsilon_{ij}^2\\
&=
\frac{1}{N}
\sum_{i=1}^{N}
e_i^\top M_{\mathcal G}e_i
=B_2^2.
\end{aligned}
\end{equation}

Since $B_1,B_2\geq0$,
this establishes
\[
0\leq
\widehat R_{\mathrm{exc}}
\leq B_1\leq B_2.
\]

These inequalities are deterministic
identities and bounds for the specified
evaluation arrays.
Their validity does not require
asymptotic approximations or
independence between evaluated goals.

Numerical verification checks
whether the implementation agrees
with these relationships to the
reported computational precision.
It does not establish an independent
population-level generalization
guarantee or a formal
floating-point error certificate.

For each configuration, define
the decomposition residual
\begin{equation}
\begin{aligned}
E_{\mathrm{tot}}
=\big|
&(\widehat R_{\mathrm{learn}}
-\widehat R_{\mathrm{full}}^\star)\\
&-\widehat R_{\mathrm{irr}}
-\widehat R_{\mathrm{exc}}
\big|.
\end{aligned}
\label{eq:app_total_regret_residual}
\end{equation}

The maximum observed
decomposition residual is below
$4\times10^{-13}$.

For CV shrinkage at $n=6912$,
the irreducible and excess
regret components sum to
$0.41205$ on DEV and
$0.49287$ on SHIFTED\_DEV,
respectively.

These checks verify the empirical
regret identities and bounds.
They do not establish a
sample-size convergence rate
or guarantee improved regret
for every goal distribution.

For a finite equally weighted
goal family
$\mathcal G=\{g_1,\ldots,g_K\}$,
define its empirical mean
and covariance by
\begin{equation}
\begin{aligned}
\bar g
&=\frac1K\sum_{j=1}^{K}g_j,\\
S_{\mathcal G}
&=\frac1K\sum_{j=1}^{K}
(g_j-\bar g)(g_j-\bar g)^\top.
\end{aligned}
\label{eq:app_goal_covariance}
\end{equation}

Since the goal feature is
$\phi(g)=(1,-2g^\top)^\top$,
the projected-error matrix satisfies
\begin{equation}
M_{\mathcal G}
=
\begin{pmatrix}
1 & -2\bar g^\top\\
-2\bar g &
4(S_{\mathcal G}+\bar g\bar g^\top)
\end{pmatrix}.
\label{eq:app_goal_matrix_blocks}
\end{equation}

The Schur complement of the
upper-left scalar block is
$4S_{\mathcal G}$.
Consequently,
\begin{equation}
\begin{aligned}
\operatorname{rank}(M_{\mathcal G})
&=1+\operatorname{rank}(S_{\mathcal G})\\
&=1+\dim\operatorname{aff}(\mathcal G).
\end{aligned}
\label{eq:app_goal_matrix_rank}
\end{equation}

Thus, a single goal induces
rank one, a goal family supported
on an affine line induces rank
two, and non-collinear planar
goals induce rank three.

In the rank-two case,
$M_{\mathcal G}$ has a
one-dimensional nullspace.
Estimation errors in this
nullspace do not contribute
to projected-contrast error
for goals on the evaluated line.

However, those errors can
affect action comparisons for
new goals outside that line.
Therefore, low projected error
on an affine-line goal family
does not guarantee reliable
decisions for novel goal
directions.

For non-collinear planar goals,
$M_{\mathcal G}$ has full rank,
but this fact alone does not
guarantee low estimation error
or reliable generalization
under distribution shift.

We next characterize which state-estimation
errors are detectable through the registered
goal family.
Let $\mathcal G=\{g_1,\ldots,g_K\}$
be the finite goal family defined above,
and let
$e_n(h)=\widehat z_n(h)-z(h)$.
For each goal, define the projected
contrast error
\[
\epsilon_{g_j}(h)
=
\langle e_n(h),\phi(g_j)\rangle.
\]

The average squared contrast error
has the exact quadratic representation
\begin{equation}
\label{eq:app_goal_error_gram}
\begin{aligned}
&\frac{1}{K}
\sum_{j=1}^{K}
|\epsilon_{g_j}(h)|^2\\
&\qquad=
e_n(h)^\top
M_{\mathcal G}e_n(h).
\end{aligned}
\end{equation}

Furthermore, the null space of
$M_{\mathcal G}$ consists exactly
of state-error directions that leave
all registered goal contrasts unchanged:
\begin{equation}
\label{eq:app_goal_error_nullspace}
\begin{aligned}
\ker M_{\mathcal G}
=
\bigl\{
u\in\mathbb R^3:
\langle u,\phi(g_j)\rangle=0\\
\text{for all }j=1,\ldots,K
\bigr\}.
\end{aligned}
\end{equation}

In particular,
\begin{equation}
\label{eq:app_goal_error_nullity}
\dim\ker M_{\mathcal G}
=
2-\dim\operatorname{aff}(\mathcal G).
\end{equation}

If $\mathcal G$ contains three
non-collinear planar goals,
then $M_{\mathcal G}$ is positive
definite and
\begin{equation}
\label{eq:app_goal_error_spectral}
\begin{aligned}
&\lambda_{\min}(M_{\mathcal G})
\|e_n(h)\|_2^2\\
&\quad\leq
\frac{1}{K}
\sum_{j=1}^{K}
|\epsilon_{g_j}(h)|^2\\
&\quad\leq
\lambda_{\max}(M_{\mathcal G})
\|e_n(h)\|_2^2.
\end{aligned}
\end{equation}

By definition of the goal-feature
Gram matrix,
\[
M_{\mathcal G}
=
\frac{1}{K}
\sum_{j=1}^{K}
\phi(g_j)\phi(g_j)^\top.
\]
Therefore,
\[
\begin{aligned}
e_n^\top M_{\mathcal G}e_n
&=
\frac{1}{K}
\sum_{j=1}^{K}
(e_n^\top\phi(g_j))^2\\
&=
\frac{1}{K}
\sum_{j=1}^{K}
|\epsilon_{g_j}|^2.
\end{aligned}
\]
This proves
\eqref{eq:app_goal_error_gram}.

Because the right-hand side is a
sum of nonnegative terms,
it vanishes precisely when
$\langle e_n,\phi(g_j)\rangle=0$
for every registered goal.
This proves
\eqref{eq:app_goal_error_nullspace}.

The rank identity established above gives
\[
\operatorname{rank}(M_{\mathcal G})
=
1+\dim\operatorname{aff}(\mathcal G).
\]
Since $M_{\mathcal G}$ is a
$3\times3$ matrix, rank-nullity yields
\eqref{eq:app_goal_error_nullity}.

For three non-collinear goals,
the affine dimension equals two.
Thus $M_{\mathcal G}$ has rank three.
As a positive-semidefinite Gram matrix,
it is therefore positive definite.
The standard Rayleigh-quotient bounds
give \eqref{eq:app_goal_error_spectral}.

For a single distinct goal,
two independent state-error directions
are invisible to its action comparison.
For goals lying on an affine line,
one such direction remains.
A non-collinear planar goal family
has no nonzero invisible direction.

These are algebraic statements about
exact goal contrasts.
They do not establish statistical
identifiability of the Hilbert state
from finite or incomplete observations.

This goal-dependent error geometry also
determines when exact contrast predictions
transfer beyond the registered goal set.
Fix an observation history $h$.
Write the Hilbert-state estimation error as
\[
e_n(h)=(e_\kappa(h),e_d(h)),
\qquad e_d(h)\in\mathbb R^2.
\]

Suppose the registered training goals
contain at least two distinct points
and have affine hull
\[
L=g_0+\operatorname{span}\{v\},
\qquad v\neq 0.
\]
Assume that the projected contrast
error vanishes at every registered goal:
\[
\epsilon_{g_j}(h)=0
\quad\text{for all }g_j\in\mathcal G.
\]

Then the projected contrast error
vanishes for every $g\in L$,
including goals outside the convex
hull of the registered goals.

For an arbitrary goal $g\in\mathbb R^2$,
\begin{equation}
\label{eq:app_affine_goal_transfer}
\begin{aligned}
\epsilon_g(h)
&=-2e_d(h)^\top
\bigl(g-P_L(g)\bigr),\\
|\epsilon_g(h)|
&\leq
2\|e_d(h)\|_2
\operatorname{dist}(g,L),
\end{aligned}
\end{equation}
where $P_L$ denotes Euclidean
projection onto the affine line $L$.

For any two distinct registered goals
$g_i,g_j$, the assumed zero
projected errors imply
\[
\begin{aligned}
0
&=\epsilon_{g_i}(h)
-\epsilon_{g_j}(h)\\
&=-2e_d(h)^\top(g_i-g_j).
\end{aligned}
\]
Since their difference spans the
direction of $L$, this gives
$e_d(h)^\top v=0$.

The projected-error function is affine
in the goal:
\[
\epsilon_g(h)
=e_\kappa(h)-2e_d(h)^\top g.
\]
Its vanishing at any registered goal
therefore implies
\[
e_\kappa(h)=2e_d(h)^\top g_0.
\]
Consequently,
\[
\epsilon_g(h)
=-2e_d(h)^\top(g-g_0).
\]

Because $e_d(h)$ is orthogonal to
the direction of $L$,
\[
e_d(h)^\top(P_L(g)-g_0)=0.
\]
Substitution proves the first identity
in \eqref{eq:app_affine_goal_transfer}.
Cauchy--Schwarz gives the second.

Consider training goals on the line
$L=\{(x,0):x\in\mathbb R\}$.
Let the true and estimated states be
\[
z(h)=(\delta,0,0),
\qquad
\widehat z_n(h)=(\delta,0,t),
\]
where $\delta>0$ and $t>0$.

Every goal on $L$ has identical
true and estimated contrasts:
\[
\widehat\Delta_{(x,0)}(h)
=\Delta_{(x,0)}(h)=\delta.
\]

For a novel goal
$\widetilde g=(0,y)$ with $y>0$,
however,
\begin{equation}
\label{eq:app_novel_direction_failure}
\begin{aligned}
\Delta_{\widetilde g}(h)&=\delta>0,\\
\widehat\Delta_{\widetilde g}(h)
&=\delta-2ty.
\end{aligned}
\end{equation}
Choosing $t>\delta/(2y)$
makes the estimated contrast negative.
The learned and observation-optimal
actions are then opposite, with
conditional excess regret $\delta$.

Thus exact contrast prediction
throughout the registered affine hull
does not guarantee correct decisions
along an unseen goal direction.
This is a structural counterexample,
not an assertion about the trained
estimators in the reported experiments.

The exact Hilbert representation preserves
the numerical action contrast, which is
stronger than preserving only the
selected action.
We formalize this distinction.

For $z=(\kappa,d)\in\mathbb R^3$,
define the affine contrast
\begin{equation}
f_z(g)
=
\kappa-2d^\top g,
\qquad g\in\mathbb R^2,
\label{eq:app_decision_equivalence_contrast}
\end{equation}
and let $\pi_z(g)$ select $a_0$
when $f_z(g)\le0$ and $a_1$
otherwise.

Let $z=(\kappa,d)$ and
$z'=(\kappa',d')$ satisfy
$d\neq0$ and $d'\neq0$.
Then
\begin{equation}
\begin{aligned}
&\pi_z(g)=\pi_{z'}(g)
\quad\text{for all }g\in\mathbb R^2\\
&\qquad\Longleftrightarrow
\quad z'=\alpha z
\quad\text{for some }\alpha>0.
\end{aligned}
\label{eq:app_global_decision_equivalence}
\end{equation}
In contrast, exact equality of
the contrasts for every goal implies
$z'=z$.

First, if $z'=\alpha z$ with $\alpha>0$,
then
$f_{z'}(g)=\alpha f_z(g)$,
so both contrasts induce the same
binary action for every goal.

Conversely, suppose the selected
actions agree for every
$g\in\mathbb R^2$.
Their positive-contrast regions
must coincide:
\begin{equation}
\{g:f_z(g)>0\}
=
\{g:f_{z'}(g)>0\}.
\label{eq:app_positive_halfspace_equality}
\end{equation}
Because $d$ and $d'$ are nonzero,
both sets are nonempty open
half-spaces.
Equality of these half-spaces
requires their boundary lines
to coincide and their orientations
to agree.
Hence their affine defining
functions differ by a positive
scalar:
\[
f_{z'}(g)
=
\alpha f_z(g),
\qquad \alpha>0.
\]
Matching the constant and linear
coefficients gives
$\kappa'=\alpha\kappa$ and
$d'=\alpha d$, proving
\eqref{eq:app_global_decision_equivalence}.

If instead
$f_{z'}(g)=f_z(g)$ for all $g$,
equality of their affine
coefficients gives $z'=z$.
Thus, numerical contrast preservation
is strictly stronger than
decision equivalence.

Agreement of selected actions at finitely
many registered goals need not extend
even to their convex hull.
Consider the triangular goal region
\[
\mathcal T
=
\operatorname{conv}
\{(0,0),(1,0),(0,1)\}
\]
and the two affine contrasts
\begin{equation}
f_z(g)=g_1-\frac12,
\qquad
f_{z'}(g)=g_1-\frac35.
\label{eq:app_finite_goal_counterexample}
\end{equation}
At all three vertices of
$\mathcal T$, these functions have
the same nonzero sign.
Consequently, their induced
actions agree at every vertex.

However, the interior goal
\[
g^\dagger
=
\left(\frac{11}{20},\frac1{10}\right)
\in\mathcal T
\]
satisfies
\begin{equation}
f_z(g^\dagger)=\frac1{20}>0,
\qquad
f_{z'}(g^\dagger)=-\frac1{20}<0.
\label{eq:app_interior_goal_disagreement}
\end{equation}
The two decision rules therefore
select opposite actions at
$g^\dagger$.
Taking $f_z$ as the true action
contrast, the resulting
conditional excess regret is
$1/20$.

This algebraic example shows that
correct actions at finitely many
registered goals need not imply
correct actions throughout their
convex hull.
By contrast, exact contrast
equality at three non-collinear
goals determines the complete
three-dimensional affine contrast
and therefore transfers to
every planar goal.

The positive-scaling characterization
concerns nonconstant affine
contrasts over the entire goal plane.
Constant contrasts admit additional
decision-equivalent states,
while restricted goal domains
may admit equivalences not described
by positive scaling.
The counterexample concerns
algebraic Hilbert states and does
not assert that the reported
trained estimators exhibit
this particular failure.

The preceding counterexample concerns
agreement of discrete actions.
Numerical contrast errors obey a
stronger geometric relation because
they are affine in the goal.

Fix a history $h$ and three
non-collinear registered goals
$g^{(1)},g^{(2)},g^{(3)}\in\mathbb R^2$.
Every planar goal $g$ admits unique
barycentric coordinates
$\lambda_1(g),\lambda_2(g),\lambda_3(g)$
satisfying
\begin{equation}
\begin{aligned}
g
&=
\sum_{j=1}^{3}\lambda_j(g)g^{(j)},\\
1
&=
\sum_{j=1}^{3}\lambda_j(g).
\end{aligned}
\label{eq:app_barycentric_goals}
\end{equation}

The affine structure of the Hilbert
contrast yields an exact interpolation
identity.
For the Hilbert contrast error
$\epsilon_g(h)=\widehat\Delta_g(h)-\Delta_g(h)$,
\begin{equation}
\epsilon_g(h)
=
\sum_{j=1}^{3}
\lambda_j(g)
\epsilon_{g^{(j)}}(h).
\label{eq:app_exact_error_interpolation}
\end{equation}
Consequently, if
\[
\max_{j\in\{1,2,3\}}
|\epsilon_{g^{(j)}}(h)|
\le\delta,
\qquad \delta\ge0,
\]
then
\begin{equation}
|\epsilon_g(h)|
\le
\delta L(g),
\qquad
L(g):=
\sum_{j=1}^{3}|\lambda_j(g)|.
\label{eq:app_sharp_goal_error_bound}
\end{equation}

To establish the identity, observe
that the projected Hilbert error
has the affine form
\[
\epsilon_g(h)
=
e_\kappa(h)-2e_d(h)^\top g.
\]
Substituting
\eqref{eq:app_barycentric_goals}
and using the fact that the
barycentric coefficients sum to one,
we obtain
\begin{equation}
\begin{aligned}
\epsilon_g(h)
&=
\sum_{j=1}^{3}\lambda_j(g)
\bigl[
e_\kappa(h)
-
2e_d(h)^\top g^{(j)}
\bigr]\\
&=
\sum_{j=1}^{3}\lambda_j(g)
\epsilon_{g^{(j)}}(h).
\end{aligned}
\label{eq:app_error_interpolation_proof}
\end{equation}
The triangle inequality immediately
gives
\eqref{eq:app_sharp_goal_error_bound}.

The amplification factor $L(g)$
is sharp over the class of
affine contrast-error functions.
Let $\mathcal A_\delta$ denote
the set of affine functions
$f:\mathbb R^2\to\mathbb R$
satisfying
\[
|f(g^{(j)})|\le\delta,
\qquad j=1,2,3.
\]
Then, for every fixed planar goal $g$,
\begin{equation}
\sup_{f\in\mathcal A_\delta}
|f(g)|
=
\delta L(g).
\label{eq:app_sharp_barycentric_factor}
\end{equation}

The upper bound follows from affine
interpolation and the triangle inequality.
For the reverse inequality,
prescribe the vertex values
\[
f(g^{(j)})
=
\delta\,
\operatorname{sgn}(\lambda_j(g)),
\]
with an arbitrary sign when
$\lambda_j(g)=0$.
Because the three registered
goals are non-collinear,
there exists a unique affine
function attaining these values.
For this function,
\[
f(g)
=
\delta
\sum_{j=1}^{3}
|\lambda_j(g)|
=
\delta L(g).
\]
Hence the upper bound is attained,
proving
\eqref{eq:app_sharp_barycentric_factor}.

The resulting error bound distinguishes
interpolation from extrapolation.
If $g$ belongs to
\[
\mathcal T
=
\operatorname{conv}
\{g^{(1)},g^{(2)},g^{(3)}\},
\]
then all its barycentric coefficients
are nonnegative.
Therefore,
\begin{equation}
g\in\mathcal T
\quad\Longrightarrow\quad
L(g)=1,
\qquad
|\epsilon_g(h)|\le\delta.
\label{eq:app_convex_hull_error_control}
\end{equation}
For $g\notin\mathcal T$,
at least one barycentric coefficient
is negative, so $L(g)>1$.
The factor $L(g)$ quantifies the
possible amplification of
registered-goal errors during
extrapolation.

In particular, if $\delta=0$,
the contrast error vanishes for
every planar goal.
For nonzero $\delta$, no uniform
error bound over the entire
unbounded goal plane follows
from the three registered-goal
errors alone.

At any target goal satisfying
\begin{equation}
|\Delta_g(h)|
>
\delta L(g),
\label{eq:app_interpolated_action_margin}
\end{equation}
the estimated and exact action
contrasts have the same sign,
and hence select the same action.

The sharpness statement holds
over the class of unrestricted
affine contrast-error functions.
It is not a claim that the
reported learned estimators
attain the worst-case bound.
Using the result as a deployment
certificate additionally requires
justified bounds on the registered-goal
contrast errors.

We evaluate the learned decision states on goal
families excluded from hyperparameter selection.

The shrinkage coefficient is
selected using training histories
and 25 registered goals only.
At $n=6912$, the selected value is
$\widehat\lambda_{\mathrm{CV}}
=0.40942928$.

We consider 25 extrapolated planar
goals and five novel-direction goals.
Table~\ref{tab:app_ood_regret}
reports the resulting excess regret.

\begin{table}[H]
\centering
\caption{
Excess regret on out-of-support
goal families at $n=6912$.
Mean, Full, and CV denote mean-only,
full-moment, and cross-validated
variance-shrinkage estimation.
}
\label{tab:app_ood_regret}
\small
\setlength{\tabcolsep}{3pt}
\begin{tabular}{@{}lccc@{}}
\toprule
Goal family & Mean & Full & CV\\
\midrule
DEV: extrapolated
& 0.13680 & 0.13807 & 0.08931\\
Shifted: novel direction
& 0.06181 & 0.20998 & 0.08709\\
\bottomrule
\end{tabular}
\end{table}

We make explicit the statistical estimand
underlying the reported paired comparisons.
Fix a regime, a goal family
$\mathcal G=\{g_1,\ldots,g_K\}$,
and the trained estimators.
For method $m$, define its
history--goal regret contribution as
\begin{equation}
\ell_{ij}^{(m)}
=
|\Delta_{g_j}(h_i)|
\mathbf 1\{
\widehat\pi_m(h_i,g_j)
\neq
\pi^\star(h_i,g_j)
\}.
\label{eq:app_paired_regret_contribution}
\end{equation}
The paired contribution of cross-validated
shrinkage relative to the mean-only
estimator is
\begin{equation}
d_{ij}
=
\ell_{ij}^{(\mathrm{CV})}
-
\ell_{ij}^{(\mathrm{mean})}.
\label{eq:app_paired_regret_difference}
\end{equation}
Both losses use the same history, goal,
and exact reference contrast.
Consequently, the comparison retains
their within-history dependence.

The evaluation contains $C=12$
registered groups per regime,
each with $128$ histories.
Let $\mathcal I_c$ denote the histories
in group $c$.
The group-level paired difference and
its empirical average are
\begin{equation}
\begin{aligned}
D_c
&=
\frac{1}{128K}
\sum_{i\in\mathcal I_c}
\sum_{j=1}^{K}d_{ij},\\
\widehat D
&=
\frac{1}{C}\sum_{c=1}^{C}D_c.
\end{aligned}
\label{eq:app_group_paired_difference}
\end{equation}
Thus, $\widehat D<0$ favors
cross-validated shrinkage,
whereas $\widehat D>0$
favors the mean-only estimator.

For bootstrap replicate $b$, draw
$C_c^{(b)}$ independently and uniformly
from $\{1,\ldots,C\}$ for
$c=1,\ldots,C$.
\mbox{The bootstrap estimate is}
\begin{equation}
\widehat D^{(b)}
=
\frac{1}{C}
\sum_{r=1}^{C}D_{C_r^{(b)}}.
\label{eq:app_group_bootstrap_replicate}
\end{equation}
The percentile interval from
$B=2000$ replicates is
\begin{equation}
\mathrm{CI}_{95\%}
=
\left[
q_{0.025}
\bigl(\{\widehat D^{(b)}\}_{b=1}^{B}\bigr),
q_{0.975}
\bigl(\{\widehat D^{(b)}\}_{b=1}^{B}\bigr)
\right].
\label{eq:app_paired_bootstrap_interval}
\end{equation}

Resampling complete groups preserves
the pairing between estimators
and the dependence among histories
within each selected group.
The resulting intervals describe
variation under this group-resampling
scheme, conditional on the evaluated
goal family and fitted estimators.
Their nominal coverage is not an
exact finite-sample guarantee,
particularly with only twelve groups.
They do not account separately
for retraining variability
or establish superiority under
arbitrary goal distributions.

For each history--goal pair $(i,j)$,
let $I_{ij}^{\mathrm{CV}}$ and
$I_{ij}^{\mathrm{mean}}$ indicate disagreement
with the exact observation-optimal action.
With $N=128C$, the group-averaged paired
excess-regret difference defined above
can equivalently be written as
\begin{equation}
\widehat D
=
\frac{1}{NK}
\sum_{i=1}^{N}
\sum_{j=1}^{K}
|\Delta_{ij}|
\left(
I_{ij}^{\mathrm{CV}}
-
I_{ij}^{\mathrm{mean}}
\right).
\end{equation}

\begin{figure}[!t]
\centering
\begin{tikzpicture}[
    x=43cm,
    y=0.78cm,
    font=\scriptsize
]

\draw[
    gray!65,
    densely dashed
]
(0,0.2) -- (0,2.65);

\node[
    anchor=west,
    font=\scriptsize
] at (-0.09,2.55)
{DEV: extrapolated planar goals};

\draw[
    blue!75!black,
    line width=1.2pt
]
(-0.08138,2.12) --
(-0.01607,2.12);

\foreach \x in {-0.08138,-0.01607}{
    \draw[
        blue!75!black,
        line width=0.8pt
    ]
    (\x,1.99) -- (\x,2.25);
}

\fill[
    blue!75!black
]
(-0.04749,2.12)
circle[radius=1.8pt];

\node[
    anchor=west,
    font=\scriptsize
] at (-0.09,1.45)
{SHIFTED\_DEV: novel-direction goals};

\draw[
    orange!85!black,
    line width=1.2pt
]
(0.00517,1.02) --
(0.04854,1.02);

\foreach \x in {0.00517,0.04854}{
    \draw[
        orange!85!black,
        line width=0.8pt
    ]
    (\x,0.89) -- (\x,1.15);
}

\fill[
    orange!85!black
]
(0.02528,1.02)
circle[radius=1.8pt];

\draw[
    ->,
    gray!75
]
(-0.09,0.12) -- (0.066,0.12);

\foreach \x/\lab in {
    -0.08/$-0.08$,
    -0.04/$-0.04$,
    0/$0$,
    0.04/$0.04$
}{
    \draw[
        gray!75
    ]
    (\x,0.05) -- (\x,0.19);

    \node[
        below,
        font=\scriptsize
    ] at (\x,0.02) {\lab};
}

\node[
    anchor=north,
    font=\scriptsize
] at (-0.012,-0.55)
{CV minus mean-only excess regret};

\end{tikzpicture}

\caption{
Paired excess-regret differences
between cross-validated variance
shrinkage and mean-only estimation
on out-of-support goal families.
Points indicate the estimated
CV-minus-mean differences;
horizontal intervals show
95\% cluster-bootstrap confidence
intervals.
Negative values favor
cross-validated shrinkage.
The two comparisons use different
evaluation regimes and goal families,
so their contrast does not isolate
the causal effect of goal geometry.
}
\label{fig:app_goal_shift_paired_regret}
\end{figure}
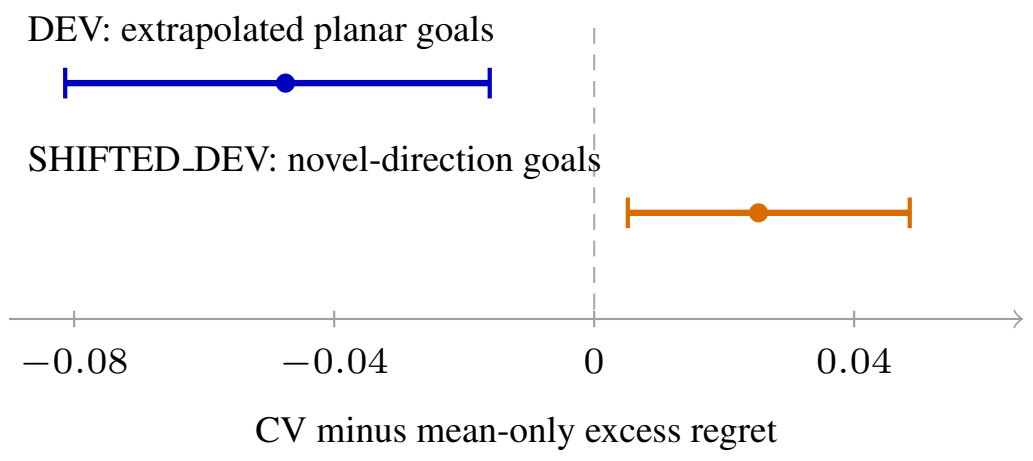

\newpage

The paired confidence intervals
are computed using 2,000
cluster-bootstrap replicates
over 12 physical groups per regime.

Each group contains 128 histories
and is defined by simulator seed,
parameter index, and observation
context.

Both estimators use identical
histories and goals within each
paired comparison.
The 95\% intervals are obtained
from the empirical 2.5\% and
97.5\% bootstrap quantiles.

On DEV with extrapolated planar
goals, CV shrinkage reduces
excess regret relative to
mean-only estimation by
$0.04749$, with paired
95\% confidence interval
$[-0.08138,-0.01607]$
for the CV-minus-mean difference.

On SHIFTED\_DEV with novel-direction
goals, CV shrinkage increases
excess regret relative to
mean-only estimation by
$0.02528$, with paired
95\% confidence interval
$[0.00517,0.04854]$.

Thus, the observed benefit of
variance shrinkage is not uniform
across goal families and
evaluation distributions.

Both evaluation regimes share the same exact
three-dimensional action-cost representation,
but the learned estimators differ in how they
estimate its conditional moments.
The observed differences concern finite-data
learning and goal-dependent evaluation,
without contradicting exact decision sufficiency
at the population level.

\section{Real-SAR Decision Learning and Evidence}
\label{app:real_sar_evidence}

We evaluate the feasibility of learning
goal-conditioned decision states from real
SAR images using the HRSID dataset.

For each eligible image, the two actions
correspond to selecting its left and right
image regions.
The action outcome $Y_a\in[0,1]^2$ is the
normalized centroid of the largest annotated
ship in the corresponding region.
Only images containing valid candidate
outcomes for both actions are retained.

The goal family is fixed to
\[
\mathcal G
=\{0.2,0.5,0.8\}^2,
\qquad |\mathcal G|=9.
\]

The cost of an action is
$\|Y_a-g\|_2^2$.
Its full-information oracle minimizes
this cost over the two candidate actions.

Each model receives a grayscale SAR image
resized to $64\times64$ pixels.
The central vertical columns 24--39
are masked to zero before inference.
Neither bounding-box annotations nor image
identifiers or source-group labels are
included in the model input.

The annotations are used to construct
supervised targets and evaluate oracle-relative
decision regret.

The registered task contains 1,981
eligible images.
The source-prefix-disjoint partition contains
1,404 training, 256 validation, and
321 test images.

The test set contains 20 source-prefix
groups and is evaluated on all nine goals.

The learned methods use matched convolutional
backbones and three training seeds:
2027, 2028, and 2029.
Training uses at most 60 epochs, with
early stopping based on validation data
and patience 12.
The test set is not used for model selection.

The compared predictors are:

\begin{itemize}
\item Hilbert-3D:
predicts the three conditional moment
contrasts defining $z(h)$.

\item Mean-only:
predicts the two action-outcome means
and compares their squared distances
to the goal.

\item Direct contrast:
predicts the nine goal-specific
action-cost contrasts directly.

\item Constant state:
uses an unconditional state estimated
from training data.
\end{itemize}

For test image $i$, goal $g_j$,
and predicted action $\widehat a_{ij}$,
the oracle-relative regret is

\begin{equation}
\begin{aligned}
r_{ij}
={}&
\|Y_{i,\widehat a_{ij}}-g_j\|_2^2\\
&-
\min_{a\in\{a_0,a_1\}}
\|Y_{i,a}-g_j\|_2^2.
\end{aligned}
\label{eq:app_real_sar_regret}
\end{equation}

We average this realized regret over
the 321 test images and nine goals,
then report the mean and standard
deviation over the three training seeds.
The group-macro metric instead equally
weights the 20 held-out filename-prefix
groups after averaging regret over images
and goals within each group. Learned-method
group-macro values are further averaged
over the three training seeds.

This metric measures performance against
the annotation-defined full-information
oracle for the registered image task.
It does not independently estimate the
population-level irreducible Bayes regret.

\begin{table}[t]
\centering
\caption{
Held-out real-SAR decision results.
Regret is averaged over 321 images
and nine goals.
Mean and standard deviation are computed
over three training seeds.
Lower regret is better.
}
\label{tab:app_real_sar_results}
\small
\setlength{\tabcolsep}{3.0pt}
\renewcommand{\arraystretch}{1.05}
\begin{tabular}{@{}lcc@{}}
\toprule
Method & Regret (mean $\pm$ std.) & Group macro \\
\midrule
Constant state
& $0.04495\pm0.00000$ & 0.04364 \\
Mean-only
& $0.03348\pm0.00994$ & 0.03347 \\
Direct contrast
& $0.02863\pm0.00058$ & 0.02979 \\
Hilbert-3D
& $0.03064\pm0.00133$ & 0.03215 \\
\bottomrule
\end{tabular}
\end{table}

The learned Hilbert state improves
oracle-relative decision regret
over the constant-state baseline,
demonstrating that the partially
observed SAR images contain usable
goal-conditioned decision information.

Direct contrast achieves the lowest
mean regret in this fixed nine-goal
experiment.
The Hilbert method instead estimates
a fixed three-dimensional state whose
action-cost contrast can be evaluated
algebraically for any planar goal.
Its exact representation property
does not require a separate learned
output for every queried goal.

We compute paired bootstrap intervals
using 20,000 replicates over the 20
held-out source-prefix groups.
Method comparisons use identical
test images and goals.
The reported differences average
over the three fixed training seeds.

The estimated Hilbert-minus-baseline
regret differences are

\begin{align*}
\text{Hilbert}-\text{Constant}
&=-0.01431,\\
&\quad[-0.01850,-0.00932],\\
\text{Hilbert}-\text{Mean}
&=-0.00283,\\
&\quad[-0.00558,\phantom{-}0.00008],\\
\text{Hilbert}-\text{Direct}
&=\phantom{-}0.00202,\\
&\quad[\phantom{-}0.00053,\phantom{-}0.00340].
\end{align*}

These nominal intervals condition on
the fixed trained models and exclude
retraining uncertainty.

\begin{figure}[t]
    \centering
    \includegraphics[width=0.98\columnwidth]
    {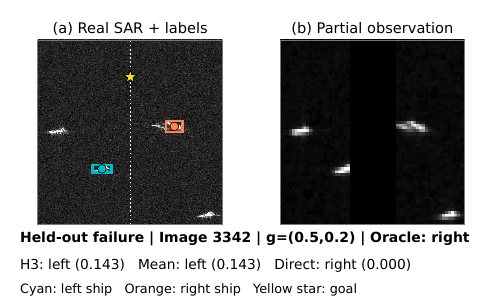}
    \caption{
        Complementary real-SAR decision evidence.
        For this held-out image and goal,
        the Hilbert-state and mean-only
        predictors select the left action,
        each incurring regret $0.143$,
        whereas the direct-contrast predictor
        selects the oracle right action.
        The example was selected post hoc
        to visualize estimator-dependent
        decision errors.
        Image contrast is adjusted for
        display only.
    }
    \label{fig:app_real_sar_failure}
\end{figure}

Figure~\ref{fig:app_real_sar_failure}
shows a complementary held-out example
in which the learned estimators make
different decisions.

Together, the two qualitative examples
illustrate how learned decision
representations operate on real SAR
observations and how finite-data
estimation affects action selection.
The aggregate results and paired
comparisons provide the quantitative
basis for evaluating reliability.

The results above establish feasibility
on the nine registered goals.
A separate evaluation of
previously unseen goals and altered
observation masks is reported in
Section~\ref{app:real_sar_ood}.


\section{Real-SAR Observation-and-Goal OOD Evaluation}
\label{app:real_sar_ood}

We further evaluate whether decision-relevant
representations learned from partially observed
real SAR images remain useful under previously
unseen goals and changes in observation geometry.
This experiment extends the registered real-SAR
task in Section~\ref{app:real_sar_evidence}
without changing its underlying action outcomes,
test partition, or annotation-defined oracle.

The evaluation uses the same 321 held-out HRSID
images and 20 filename-prefix source groups.
For each image, the two actions select the
left and right image regions, respectively.
The candidate outcomes are the normalized
centroids of the corresponding annotated ships.
The action cost remains the squared Euclidean
distance between the selected outcome and
the queried goal.

The nine registered goals are
$\mathcal G_{\mathrm{reg}}
=\{0.20,0.50,0.80\}^2$.
The additional evaluation uses eight goals
that were not included in the registered
training goal family:
\begin{equation}
\begin{aligned}
\mathcal G_{\mathrm{OOD}}=\{&
(0.05,0.05),(0.05,0.50),\\
&(0.05,0.95),(0.50,0.05),\\
&(0.50,0.95),(0.95,0.05),\\
&(0.95,0.50),(0.95,0.95)\}.
\end{aligned}
\label{eq:app_real_sar_ood_goals}
\end{equation}
Every evaluated OOD goal lies outside
the convex hull of the registered
goal coordinates.

All images are converted to grayscale,
resized to $64\times64$ pixels using
PIL bilinear interpolation, and normalized
by dividing pixel intensities by $255$.
Observation masks are then applied by
setting the designated pixels to zero.
The underlying SAR image, candidate
action outcomes, and oracle decision
costs remain fixed across mask regimes.

Table~\ref{tab:app_real_sar_ood_masks}
specifies the five observation regimes.
Pixel coordinates use zero-based indexing,
with $x$ denoting columns and $y$ denoting
rows; all intervals are half-open.
The registered observation corresponds
to the central vertical mask of width 16.

\begin{table}[t]
\centering
\caption{
Observation-mask definitions
for $64\times64$ SAR images.
All pixels in the indicated rectangle
are set to zero; other pixels remain unchanged.
}
\label{tab:app_real_sar_ood_masks}
\small
\setlength{\tabcolsep}{3pt}
\renewcommand{\arraystretch}{1.06}
\begin{tabular}{@{}lcc@{}}
\toprule
Regime & Columns $x$ & Rows $y$ \\
\midrule
Center vertical, 16 & $[24,40)$ & $[0,64)$ \\
Left vertical, 16 & $[8,24)$ & $[0,64)$ \\
Right vertical, 16 & $[40,56)$ & $[0,64)$ \\
Horizontal, 16 & $[0,64)$ & $[24,40)$ \\
Center vertical, 24 & $[20,44)$ & $[0,64)$ \\
\bottomrule
\end{tabular}
\end{table}

The central vertical width-16 regime
provides a goal-OOD evaluation under
the registered observation mechanism.
The remaining four regimes combine
unseen goals with an altered observation
mask, thereby testing joint
observation-and-goal shift.

The evaluated methods are the constant-state
baseline, Hilbert-3D, mean-only estimation,
and direct nine-goal contrast prediction.
The learned predictors retain their three
original training seeds, 2027, 2028,
and 2029. No model is retrained or selected
using the joint-OOD evaluation data.

To query the unregistered goals, the
evaluation protocol uses a least-squares
affine extension of the nine predicted
action contrasts available at the
registered goals. In particular, the
goal dependence is represented by
$\psi(g)=(1,g_1,g_2)^\top$, and the
affine coefficients are obtained by
fitting the predicted contrasts
over $\mathcal G_{\mathrm{reg}}$.
The resulting affine function is then
evaluated at each $g\in\mathcal G_{\mathrm{OOD}}$.

This extension uses predicted contrasts,
not ground-truth contrasts, for fitting.
It is an evaluation-side extension of
predictions rather than
goal-specific network retraining.
The joint-OOD evaluation reuses the
corresponding inference records;
no additional network inference is
performed in the final joint-OOD
aggregation stage.

For each observation regime, we report
oracle-relative realized regret,
action disagreement, and contrast
prediction error. Regret is computed
using the same annotation-defined
full-information oracle as in
Eq.~\eqref{eq:app_real_sar_regret}.
Each learned method is evaluated
over 321 images, eight goals,
and three fixed training seeds.
The constant-state baseline uses
its single registered predictor.

\begin{table*}[t]
\centering
\caption{
Real-SAR evaluation on eight unregistered
goals under five observation conditions.
Entries are mean oracle-relative regrets;
lower is better.
The learned methods are averaged over
three training seeds.
The first row is goal OOD under the
registered observation mask;
the remaining rows additionally change
the observation geometry.
}
\label{tab:app_real_sar_joint_ood}
\small
\setlength{\tabcolsep}{8pt}
\renewcommand{\arraystretch}{1.10}
\begin{tabular}{@{}lrrrr@{}}
\toprule
Observation regime
& Constant state
& Hilbert-3D
& Mean-only
& Direct contrast \\
\midrule
Center vertical, 16
& 0.05997 & 0.04025 & 0.04296 & 0.03694 \\
Right vertical, 16
& 0.05997 & 0.05611 & 0.06155 & 0.06490 \\
Left vertical, 16
& 0.05997 & 0.08501 & 0.06494 & 0.09487 \\
Horizontal, 16
& 0.05997 & 0.07563 & 0.06347 & 0.08440 \\
Center vertical, 24
& 0.05997 & 0.04391 & 0.04779 & 0.04167 \\
\bottomrule
\end{tabular}
\end{table*}

The results in
Table~\ref{tab:app_real_sar_joint_ood}
show that the relative performance
of learned decision representations
depends on the observation regime.
Under the registered central mask,
Hilbert-3D achieves regret $0.04025$,
compared with $0.04296$ for mean-only
estimation and $0.03694$ for direct
contrast prediction.

Under the rightward mask shift,
Hilbert-3D achieves regret $0.05611$,
below the $0.06155$ of mean-only
estimation and $0.06490$ of direct
contrast prediction.
Conversely, under the leftward shift,
Hilbert-3D regret rises to $0.08501$,
compared with $0.06494$ for mean-only
estimation.
The horizontal-mask regime also
favors mean-only estimation over
Hilbert-3D, whereas the wider central
vertical mask favors Hilbert-3D
over mean-only estimation.

These differences are evaluated
using paired source-prefix-group
bootstrap resampling.
The procedure samples the 20 held-out
filename-prefix groups with replacement,
retaining the paired evaluations
of the compared methods within each
resampled group.
The registered computation uses
2,000 bootstrap replicates with
random seed 20271111.
The intervals are empirical percentile
intervals under this exploratory
group-resampling scheme.

For the Hilbert-minus-mean-only
regret difference, the estimated
difference under the rightward shift
is $-0.00545$, with a nominal
95\% interval of
$[-0.00962,-0.00097]$.
Under the leftward shift, the difference
is $0.02008$, with interval
$[0.01066,0.03064]$.
The corresponding differences are
$0.01216$ with interval
$[0.00589,0.01853]$ for the
horizontal mask, and $-0.00388$
with interval $[-0.00616,-0.00111]$
for the wider central mask.

These intervals characterize variation
over the evaluated filename-prefix
groups, conditional on the fixed
trained predictors and goal family.
Filename-prefix groups have not
been verified as independent SAR
acquisitions. The reported intervals
therefore do not establish
acquisition-level statistical
independence or provide exact
population-level coverage guarantees.

Figure 5 of the main paper provides
two complementary visual witnesses
selected from the held-out SAR images.
Both are evaluated at the previously
unregistered goal $g=(0.50,0.95)$.

The rightward-shift witness is image
2263, corresponding to
\texttt{P0040\_5060\_5860\_6000\_6800.jpg}
from source-prefix group
\texttt{P0040}.
Its oracle action is $a_1$, with
true action-cost contrast $0.58752$.
Under the rightward observation shift,
Hilbert-3D selects the oracle action
for all three training seeds, whereas
mean-only estimation is correct for
two of the three seeds.
The respective seed-averaged regrets
are $0$ and $0.19584$.

The leftward-shift witness is image
2598, corresponding to
\texttt{P0049\_0\_800\_4200\_5000.jpg}
from source-prefix group
\texttt{P0049}.
Its oracle action is $a_0$, with
true contrast $-0.68476$.
Under the registered central mask,
Hilbert-3D selects the oracle action
for all three seeds.
Under the leftward shift,
Hilbert-3D is correct for only
two seeds, giving mean regret
$0.22825$, whereas mean-only
estimation is correct for all
three seeds and incurs zero regret.

The two witnesses were selected
after examining the evaluation
records, using a fixed
75th-percentile rank within the
respective advantage and disadvantage
candidate sets and preferring distinct
images and source-prefix groups.
The rightward and leftward candidate
sets contained 473 and 629 cases,
respectively.
These examples illustrate possible
decision outcomes under observation
shift; their selection does not
estimate how frequently such
outcomes occur in the test population.

The joint-OOD results distinguish
two properties that should not be
conflated. The exact affine
action-cost representation establishes
decision sufficiency for the specified
quadratic two-action problem when
the required conditional quantities
are known. The observed behavior
of learned estimators additionally
depends on finite training data,
estimation error, and the observation
mechanism.

Accordingly, the real-SAR experiments
support feasible goal-conditioned
decision-making from incomplete
image observations, including
queries outside the registered
goal family. They do not establish
uniform superiority of Hilbert-3D
under observation shift, identify
the population Bayes regret,
or demonstrate causal control of
an evolving SAR-imaged system.

\section{Real-Robot Outcome Forecasting}
\label{app:real_robot_forecasting}

\begin{figure*}[!b]
    \centering
    \includegraphics[width=\textwidth]{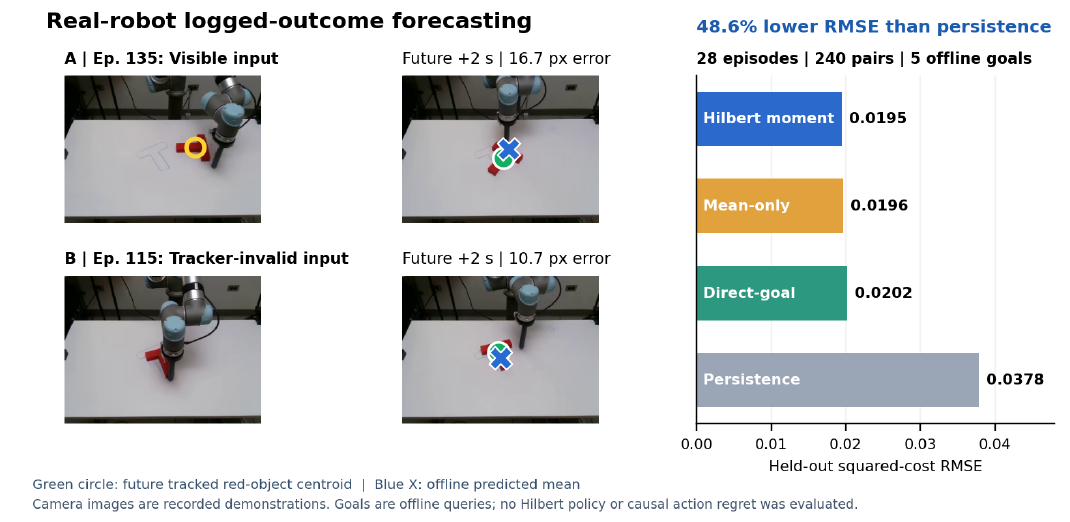}
    \caption{
        Goal-conditioned outcome forecasting on real robots.
        Left: Held-out robot observations and subsequent
        object positions after a two-second horizon.
        Green circles indicate future object centroids,
        and blue crosses indicate predictions.
        Right: Goal-cost RMSE over 240 held-out pairs
        and five offline goals.
        Hilbert moments reduce RMSE by $48.6\%$
        relative to persistence.
        No Hilbert policy was deployed;
        causal action regret is not identified.
    }
    \label{fig:app_real_robot_forecasting}
\end{figure*}

We evaluate 240 held-out input--outcome pairs
from 28 episode-disjoint test robot-push
demonstrations, following training on 108
demonstrations, across five offline image-plane
goals. Hilbert moments achieve goal-cost RMSE
$0.0195$, compared with $0.03781$ for persistence
($48.6\%$ lower). The paired episode-bootstrap
MSE-improvement interval is
$[0.000622,0.001545]$; future-object centroid
RMSE decreases from $28.72$ to $17.52$ pixels.

One moment model serves all five offline goals
without retraining. Differences from mean-only
($0.01961$) and direct-goal ($0.02017$) are
statistically unresolved. Logged actions vary
during prediction, so these forecasts do not
establish causal regret or deployed-policy
performance.

\section{Reproducibility and Evidence Scope}
\label{app:reproducibility_scope}

A separate review-stage archive contains frozen code,
nine checkpoints, reference outputs, and
integrity checks. Replay verifies C1/C2,
120 C3 checks ($n=6912$), 25,680 HF15A
rows, and HF15B (321 HRSID images,
five masks, 144,450 predictions,
20 result rows). HRSID images are obtained
separately and fingerprint-verified;
no retraining occurs. The C1 complementary
branch is analytic. The 144 smaller-sample
configurations and MuJoCo/real-robot
experiments are not independently rerun.
Replay verifies frozen-reference agreement,
not complete replication.

\end{document}